%% file: iclr2027_conference.tex
\documentclass{article} % For LaTeX2e
\usepackage{iclr2027_conference,times}

\input{math_commands.tex}

\usepackage[T1]{fontenc}
\usepackage[utf8]{inputenc}

\usepackage{hyperref}
\usepackage{url}
\usepackage{amsmath,amssymb,mathtools}
\usepackage{booktabs}
\usepackage{graphicx}
\usepackage{microtype}
\usepackage{multirow}
\usepackage[table]{xcolor}
\usepackage[skins]{tcolorbox}
\usepackage{enumitem,needspace,ragged2e}

\definecolor{FindingAccent1}{HTML}{315ED1}
\definecolor{FindingAccent2}{HTML}{137A65}
\definecolor{FindingAccent3}{HTML}{A66B12}
\definecolor{FindingAccent4}{HTML}{B64760}

\definecolor{FindingTint1}{HTML}{F3F6FC}
\definecolor{FindingTint2}{HTML}{F2F8F5}
\definecolor{FindingTint3}{HTML}{FFFBED}
\definecolor{FindingTint4}{HTML}{FBF3F5}

\definecolor{FindingRule}{HTML}{333333}
\definecolor{oursblue}{RGB}{235,243,250}
\definecolor{fullgray}{RGB}{242,242,242}
\definecolor{reductiongreen}{RGB}{35,125,80}

\newcommand{\reduction}[1]{%
  \textsubscript{
    \textcolor{reductiongreen}{\ensuremath{\downarrow}#1\%}
  }%
}

\newcommand{\findingnumber}[1]{%
  \ifcase#1\or I\or II\or III\or IV\fi
}

\newcommand{\findingbox}[2]{%
  \begin{tcolorbox}[
    enhanced,
    colback=FindingTint#1,
    colframe=FindingRule,
    boxrule=0.5pt,
    arc=4pt,
    outer arc=4pt,
    boxsep=0pt,
    left=8pt,
    right=8pt,
    top=5pt,
    bottom=5pt,
    before skip=6pt,
    after skip=4pt,
    fontupper=\normalfont\normalsize\itshape,
    before upper={
      \RaggedRight
      \hyphenpenalty=10000
      \setlength{\parindent}{0pt}
      \setlength{\parskip}{0pt}
    }
  ]
  {\normalfont\bfseries Finding~\findingnumber{#1}:}\enspace#2
  \end{tcolorbox}%
}

\setlist[itemize]{
  leftmargin=*,
  itemsep=3pt,
  topsep=4pt
}

\title{Does the VGGT Family Need All Its Layers?}

\author{
Fengyi Zhang$^{1}$,
Holger Caesar$^{2}$,
Xiangyu Sun$^{1}$,
Zheng Zhang$^{3}$,
Zi Huang$^{1}$,
Yadan Luo$^{1}$ \\
$^{1}$The University of Queensland \\
$^{2}$Delft University of Technology \\
$^{3}$Harbin Institute of Technology
}

\iclrfinalcopy

\begin{document}
\vspace*{-0.45in}
\maketitle

\fancyhead[L]{Preprint}

\begin{abstract}
Which layers of a feed-forward geometry model are needed to preserve both camera poses and dense 3D structure? We study layer redundancy in VGGT, $\pi^3$, and VGGT-$\Omega$: 3{,}018 pruned configurations, scored on seven camera-pose and dense-geometry metrics across four indoor and outdoor datasets. Four findings follow: (i) Removable layers cluster in two redundancy regions: a dominant early region and a narrower late one, while deletions spanning the intervening layers are consistently more disruptive. This recurring pattern holds across models, datasets, and metrics, and contrasts with the middle-to-late redundancy commonly reported in the literature. (ii) Within these regions, we observe that the joint degradation from deleting two intervals is approximately the sum of their individual degradations, reducing the number of model evaluations for pruning search from $O(L^4)$ to $O(L^2)$, where $L$ is the aggregator depth. (iii) We find that CKA provides a cheaper representation-based proxy for interval degradation, offering a practical trade-off between pruning quality and calibration cost. (iv) Closed-form linear calibration recovers accuracy after pruning without end-to-end retraining. A least-squares analysis shows that using a shared map for special and patch tokens generally incurs excess reconstruction loss, motivating token-aware recovery. Recovery maps fitted on just 100 calibration scenes generalize to held-out scenes and unseen datasets. The resulting models reduce aggregator parameters by up to 44\% while maintaining accuracy comparable to their intact counterparts.
Code and experimental results will be available at our
\href{https://xian-bei.github.io/vggt-family-layer-redundancy/}{\underline{project page}}.

\end{abstract}

\vspace{-1em}
\section{Introduction}
\vspace{-0.5em}
\label{sec:intro}

Feed-forward geometry models jointly recover camera poses and dense 3D structure from multi-view images in a single forward pass. VGGT \citep{vggt} and its successors, including $\pi^3$ \citep{pi3} and VGGT-$\Omega$ \citep{vggtomega} (collectively, the \textit{VGGT family}), have established a common framework for this task, demonstrating performance gains from scaling backbone depth and width. Yet, these benefits do not necessarily reveal \textit{how much of a trained model's depth is actually needed at inference}. Understanding layer redundancy can therefore inform the design of shallower and more efficient models and complement existing acceleration methods such as token pruning or merging \citep{fastvggt} and post-training quantization \citep{quantvggt}.

Studies of layer redundancy in LLMs often identify removable computation in \textit{middle-to-late} layers~\citep{uidl,curseofdepth,whatmatters}, although the inferred pattern can change substantially with the calibration criterion~\citep{rethinkingredundancy}. Analyses of some ViTs also report feature stabilization in deeper blocks~\citep{deepvit,simpler}, while cross-layer representation patterns vary with architecture and pretraining strategy~\citep{howvitworks,darksecrets,teachingmatters}. These findings provide useful starting points but offer no universal prescription for where to remove layers. This motivates us to ask: \textit{Does the VGGT family exhibit consistent patterns of layer redundancy across models, datasets, and metrics, and if so, how can these patterns guide layer pruning and post-pruning recovery?}

\begin{figure}[t]
% \vspace{-1ex}
    \centering
    \includegraphics[width=\linewidth]{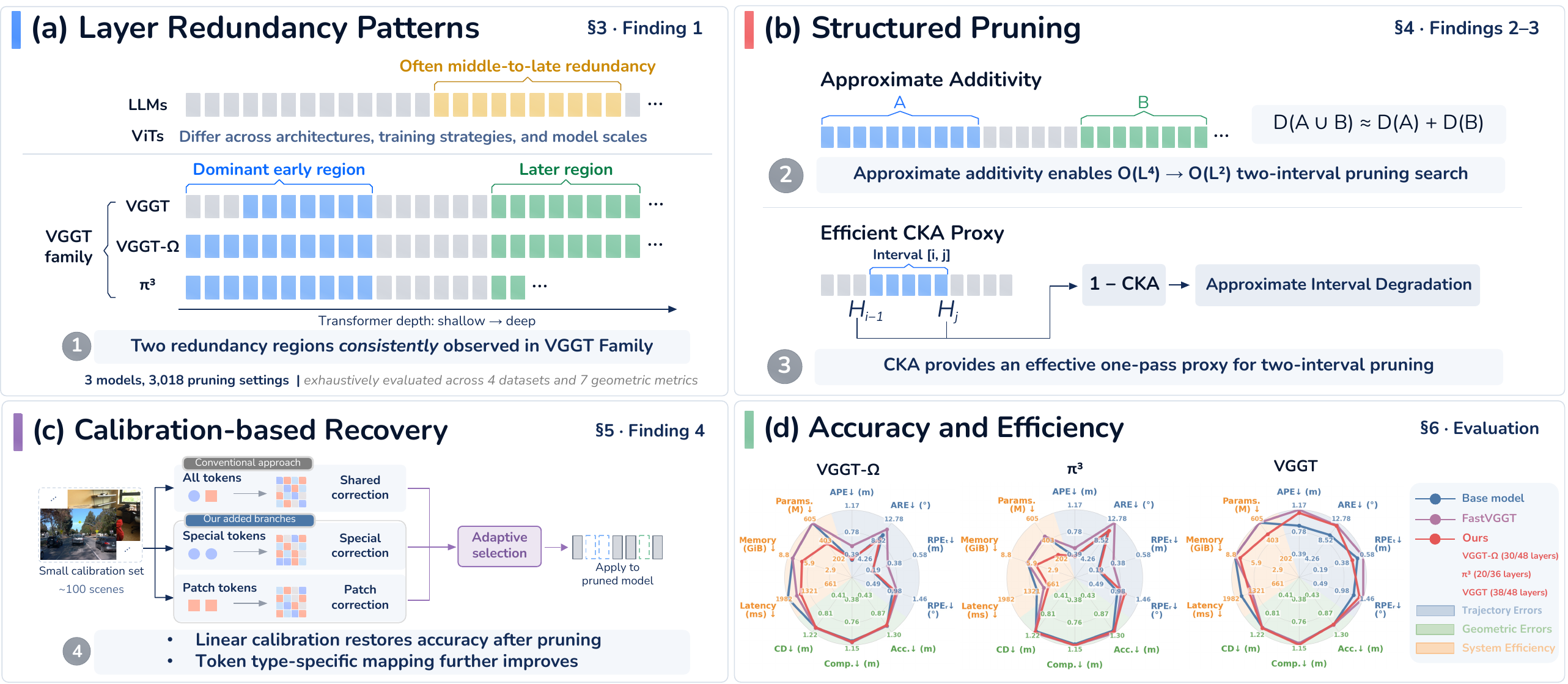}
    \vspace{-2ex}
    \caption{\textbf{Overview of our analysis and pruning framework.}
(a) Layer-redundancy patterns in the VGGT family.
(b) Structured pruning based on approximate additivity and CKA.
(c) Calibration-based recovery after pruning.
(d) Accuracy and efficiency of the resulting compressed models.}
    \label{fig:teaser}
    \vspace{-1em}
\end{figure}

We systematically study redundancy by removing contiguous intervals from the multi-view aggregator, which accounts for the largest share of parameters in each model.
For an aggregator with $L$ layers, we evaluate all $\binom{L+1}{2}$ single-interval deletions, covering 3{,}018 configurations across the three backbones.
We measure their effects on camera trajectories and dense reconstruction across four indoor and outdoor datasets.
The resulting \emph{interval-deletion landscapes} reveal where layers can be removed with negligible loss of accuracy and provide the basis for studying joint deletions, efficient pruning selection, and calibration-based recovery.
Our analysis yields four findings, summarized in Fig.~\ref{fig:teaser}:

\begin{itemize}
    \item The VGGT family exhibits two separated redundancy regions: a broad early region and a narrower late one.
    This structure is remarkably consistent across models, datasets, and evaluation metrics, motivating a restriction of the pruning search space from arbitrary layer combinations to configurations comprising up to two separated intervals.

    \item Degradation is approximately additive for deletions within the two redundancy regions.
    Estimating joint degradation by summing individual-interval degradations reduces the required model evaluations for pruning search from $O(L^4)$ to $O(L^2)$.
    The resulting two-interval strategy preserves accuracy better than single-interval pruning and the compared methods in nearly all settings.

    \item Centered Kernel Alignment (CKA) \citep{cka} provides a practical proxy for pruning search using one intact-model forward pass per calibration input.
    It generally performs best among the compared proxies, offering a practical trade-off between calibration cost and pruning quality.

    \item Closed-form linear calibration effectively recovers accuracy after pruning without end-to-end training.
    A least-squares analysis reveals that special and patch tokens generally benefit from separate corrections, motivating token-aware calibration, which provides further accuracy gains.
\end{itemize}
\vspace{-1ex}
Using only 100 calibration scenes without end-to-end retraining, the resulting models maintain accuracy comparable to their intact counterparts while removing up to 44\% of aggregator layers, substantially reducing parameter counts and inference latency.
The approach generalizes well across scenes and sequence lengths and remains complementary to token reduction and quantization.

\vspace{-0.7em}
\section{Probing Layer Redundancy in the VGGT Family}
\label{sec:setup}

\subsection{Models and layer redundancy}\vspace{-1ex}

We study the multi-view \emph{aggregators} of pretrained VGGT~\citep{vggt}, $\pi^3$~\citep{pi3}, and VGGT-$\Omega$~\citep{vggtomega}, which account for the largest share of parameters in each model. A \emph{layer} is one complete transformer block, including its attention and feed-forward sublayers; within-view and cross-view blocks are counted separately. \emph{Depth} denotes the number of aggregator layers, with $L=48$, $36$, and $48$ for VGGT, $\pi^3$, and VGGT-$\Omega$, respectively.
We assess \emph{layer redundancy} through the effect of layer removal on geometric prediction accuracy under a specified evaluation setting. 
For an intact model $F$, $F_{-I}$ denotes the model obtained by bypassing the blocks indexed by $I\subseteq\{1,\ldots,L\}$ while retaining the image encoder and prediction heads. A contiguous interval of layers is denoted by $I=[i,j]=\{i,\ldots,j\}$, and the pruning budget is $b=|I|$. 

\subsection{Datasets and evaluation metrics}\vspace{-1ex}

We use 30 frames per scene from the \emph{indoor} datasets ScanNet~\citep{scannet} and 7Scenes~\citep{7scenes} and the \emph{outdoor} datasets nuScenes~\citep{nuscenes} and Waymo~\citep{waymo}. We use disjoint calibration and evaluation splits of 50 scenes each, except for 7Scenes, where we use the official splits.
Trajectories are evaluated using absolute position and rotation errors (APE, ARE) and relative translation and rotation errors ($\mathrm{RPE}_t$, $\mathrm{RPE}_r$); dense reconstruction is evaluated using accuracy (Acc.), completeness (Comp.), and Chamfer distance (CD). All metrics are \emph{lower-is-better}, with translation and reconstruction errors measured in metres and rotation errors in degrees.

\subsection{Measuring geometric degradation}\vspace{-1ex}

For a fixed model and dataset, let $P_m(\cdot)$ denote the error under metric $m$, averaged over the scenes in the relevant split. We define the degradation caused by deleting $I$ as $D_m(I)=P_m(F_{-I})-P_m(F)$, where positive values indicate degradation. This quantity measures a change in error relative to ground truth, rather than a representation distance or a deviation from the intact model's outputs.
Because the seven metrics differ by orders of magnitude, we normalize their degradation before aggregating them into \emph{a single scalar score} that measures the degradation caused by deleting $I$.
For a set of evaluation metrics $\mathcal M$, we define
$D_{\mathcal M}(I)=|\mathcal M|^{-1}\sum_{m\in\mathcal M}D_m(I)/\alpha_m$,
where $|\mathcal M|$ is the number of metrics and $\alpha_m$ is the largest absolute degradation over all single-interval deletions.
For each model and dataset, these normalization scales are computed on the calibration split and held fixed across pruning budgets and evaluation splits.
When multiple datasets are considered, we use $D_{\mathcal M}$ to denote the equally weighted average of their individually normalized scores.
Appendix~\ref{app:raw_results} reports the unnormalized raw degradation for all seven metrics.

\vspace{-0.3em}
\section{Where Are Layers Redundant?}\vspace{-1ex}
\label{sec:redundancy}
For each model and dataset, we evaluate all $\binom{L+1}{2}$ single-interval deletions on the calibration split.
In Fig. \ref{fig:redundancy_heatmap}, each entry of the resulting \emph{interval-deletion landscape} records $D_{\mathcal M}([i,j])$, the normalized degradation after bypassing blocks $i$ through $j$.
Each panel averages normalized degradation equally across its two datasets.
Indexing by both endpoints distinguishes removal location from interval length: intervals with a fixed pruning budget $b$ form a diagonal slice, $D_{\mathcal M}([s,s+b-1])$.

\subsection{Two redundancy regions of different sizes}
\begin{figure}[t]
    \centering
    \includegraphics[width=0.98\linewidth]{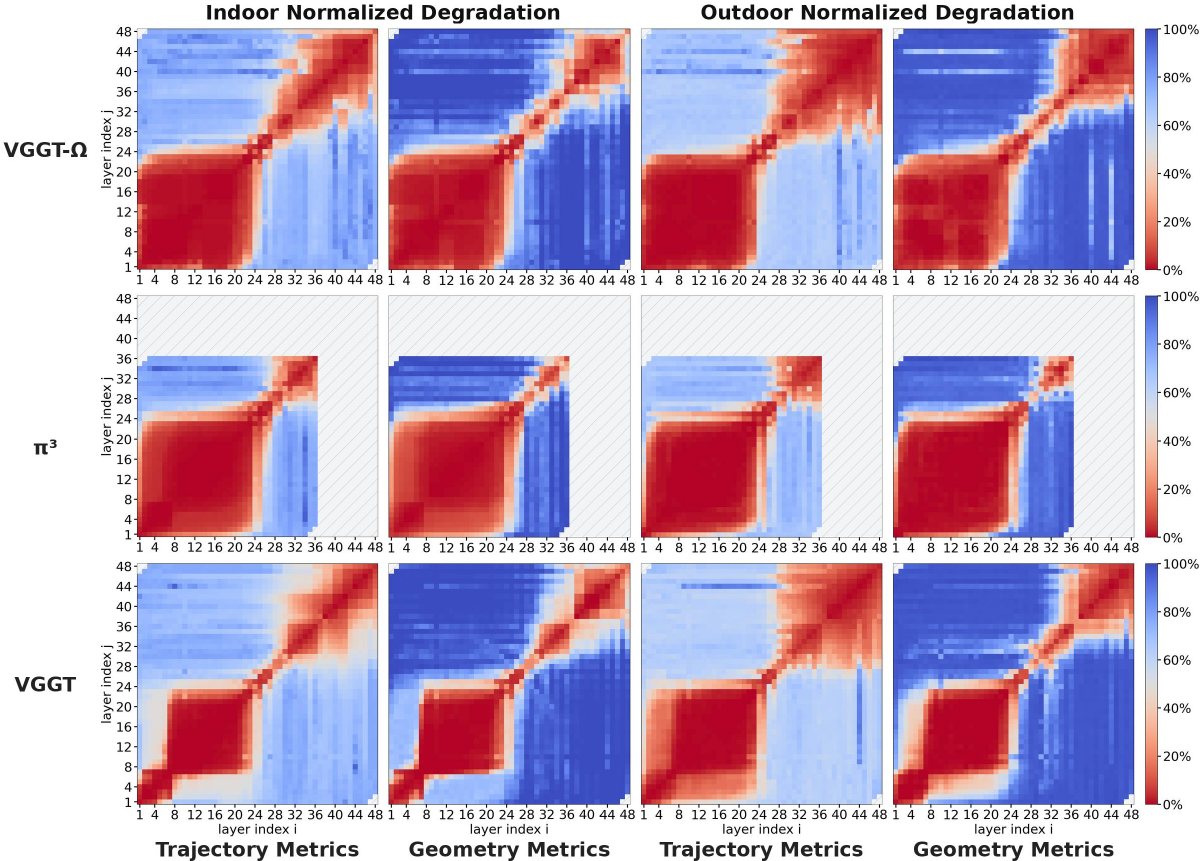}
    \vspace{-0.5em}
    \caption{
    \textbf{Two separated low-degradation regions consistently appear across models, domains and metrics, with the dominant region occurring early in the aggregator.}
    Each cell reports normalized degradation after deleting the contiguous interval $[i,j]$ of aggregator layers.
    % , before post-pruning recovery.
    Scores are mirrored across the diagonal, with $\pi^3$ aligned to the first 36 layers and the remaining 12 positions hatched.
    Trajectory and geometry panels average over their four and three metrics, respectively.
    Indoor and outdoor panels average equally over ScanNet/7Scenes and nuScenes/Waymo, respectively.
    }
    \label{fig:redundancy_heatmap}
    \vspace{-1em}
\end{figure}

\vspace{-1ex}
Fig. \ref{fig:redundancy_heatmap} shows two separated regions of low degradation: a broad early region and a narrower late region.
Intervals that span the intervening blocks are consistently more disruptive.
The early region tolerates long contiguous removals, forming a broad square of low degradation in the mirrored heatmap.
The late region forms a narrower band near the diagonal, indicating greater sensitivity to the number of consecutively removed layers.
These observations suggest that a substantial portion of early aggregation after the image encoder can be skipped, whereas fewer consecutive layers can be removed near the output.
The dominance of early redundancy contrasts with the middle-to-late redundancy often reported in LLMs~\citep{shortgpt,uidl,curseofdepth,whatmatters} and the high similarity between deep-layer representations observed in some ViT studies~\citep{deepvit,simpler}.
The two-region structure motivates restricting the layer-pruning search space from arbitrary layer combinations to configurations comprising up to two separated intervals.
The joint effect of two intervals is examined in \S~\ref{sec:pruning}.

\subsection{Shared structure across models, datasets, and metrics}\vspace{-1ex}
The two-region structure is shared across all three models and remains stable across indoor and outdoor datasets and trajectory and geometry metrics, despite substantial architectural and training differences within the VGGT family.
Relative to VGGT, $\pi^3$ removes camera and register tokens, shortens the aggregator, and uses two-stage training~\citep{pi3}; VGGT-$\Omega$ introduces register attention, replaces DINOv2~\citep{dinov2} with DINOv3~\citep{dinov3} in the image encoder, and incorporates large-scale unlabeled data into its three-stage training~\citep{vggtomega}.
Only minor variations appear within this shared pattern.
The early redundancy region is slightly shorter in VGGT than in $\pi^3$ and VGGT-$\Omega$, while the late region is slightly broader under trajectory metrics than under geometry metrics.
Despite these differences, all three models retain the overall two-region pattern, with similar locations and relative strengths.
This consistency contrasts with the dependence of LLM redundancy on calibration objectives~\citep{rethinkingredundancy} and with variations in ViT representation similarity across architectures~\citep{howvitworks}, pretraining strategies~\citep{darksecrets}, and model scales~\citep{teachingmatters}.
It supports structured pruning across scenes and datasets without implying that redundancy is independent of data or evaluation objectives.

\findingbox{1}{The VGGT family exhibits two separated redundancy regions in the aggregator: a broad early region and a narrower late region. This structure is remarkably \textit{consistent} across models, datasets, and evaluation metrics.}

\vspace{-0.3em}
\section{How Can the Two-Region Structure Guide Pruning?}
\label{sec:pruning}\vspace{-2ex}

The observed two-region structure motivates restricting the layer-pruning search space from arbitrary layer combinations to configurations comprising up to two separated contiguous intervals.
Even under this restriction, exhaustive evaluation remains expensive.
Across all pruning budgets, there are $\binom{L+1}{2}$ single-interval configurations and $\binom{L+1}{4}$ two-interval configurations with at least one retained layer between the intervals.
This gives $\binom{L+1}{2}+\binom{L+1}{4}=O(L^4)$ unique non-empty configurations, or 213,052 candidates for $L=48$.
At an illustrative cost of one second per candidate per scene for inference and evaluation, exhaustive evaluation would require approximately 5,920 GPU hours just for a 100-scene calibration set.
We therefore approximate candidate degradation to efficiently explore the performance--compression frontier across budgets.
\vspace{-1ex}
\subsection{Approximate Additivity of Interval-Wise Degradation}
\begin{figure}[t]
    \centering
    \includegraphics[width=\linewidth]{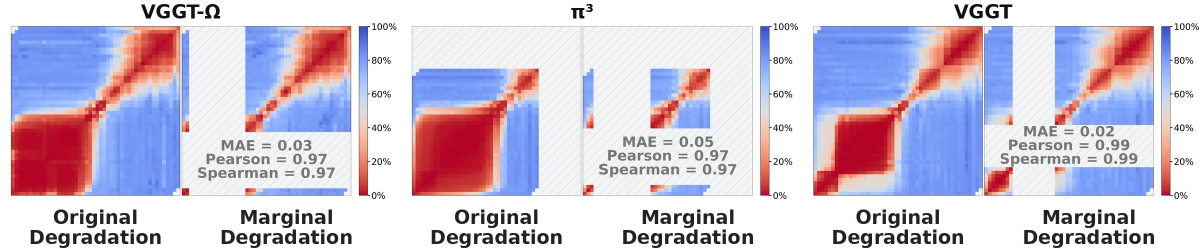}
    \vspace{-1em}
    \caption{\textbf{Approximate additivity of interval-wise degradation.}
    For each backbone, the intact-model landscape is compared with the marginal landscape after removing an interval $A$ within the early redundancy region.
    The marginal normalized degradation $D_{\mathcal M}(B\mid A)$ closely matches the original $D_{\mathcal M}(B)$ in both values and rankings. 
    }
    \label{fig:marginal_heatmap}
    \vspace{-1em}
\end{figure}

\vspace{-1ex}
\label{sec:composition}
Let $A$ and $B$ denote two separated intervals.
The marginal raw degradation measures the incremental effect of removing $B$ after $A$ has been removed:
$D_m(B\mid A)=P_m(F_{-(A\cup B)})-P_m(F_{-A})$.
Under the same normalization and aggregation, $D_{\mathcal M}(A\cup B)=D_{\mathcal M}(A)+D_{\mathcal M}(B\mid A)$ holds exactly.
We observe approximate additivity: $D_{\mathcal M}(B\mid A)\approx D_{\mathcal M}(B)$, meaning that after removing $A$, the marginal degradation landscape for $B$ remains close to the original landscape.
Fig.~\ref{fig:marginal_heatmap} illustrates this agreement using scores averaged over all four datasets: mean absolute error is 0.02--0.05, with Pearson and Spearman correlations of at least $0.97$, indicating close agreement in both values and candidate rankings.
We examine when this approximation holds in Appendix~\ref{app:additivity}.

This approximation lets us estimate joint degradation as $\widetilde D_{\mathcal M}(A\cup B)=D_{\mathcal M}(A)+D_{\mathcal M}(B)$ using only single-interval results.
It reduces the required model evaluations for two-interval pruning search from $O(L^4)$ to $O(L^2)$, while two-interval candidates are scored without additional model evaluations.
For $L=48$, the number of candidate evaluations falls from 213,052 to 1,176, reducing the illustrative cost for 100 calibration scenes from approximately 5,920 hours to less than 33 hours.

We next evaluate whether this approximation supports effective pruning across budgets.
For budget $b$, let $\mathcal C_b$ contain all configurations of one or two separated intervals removing exactly $b$ layers.
For each backbone, we use only the ScanNet and nuScenes calibration sets to select $\widetilde I_b=\arg\min_{I\in\mathcal C_b}\widetilde D_{\mathcal M}(I)$, using measured degradation for single intervals and the additive estimate for two intervals.
We compare this selection with a search restricted to single intervals.
Each configuration is then applied unchanged to their disjoint held-out sets for \emph{same-dataset generalization} and the unseen datasets 7Scenes and Waymo for \emph{cross-dataset generalization}.
We also compare against ShortGPT~\citep{shortgpt}, Gardener~\citep{gardener}, ReplaceMe~\citep{replaceme}, and SIMPLER~\citep{simpler}, with implementation details provided in Appendix~\ref{app:Pruning}.

Fig.~\ref{fig:pruning_curves} shows that allowing two intervals generally yields lower degradation than single-interval selection at larger budgets, when selected configurations begin to draw on both redundancy regions.
Gains are strongest for VGGT and VGGT-$\Omega$, whose second redundancy regions are larger, and more modest for $\pi^3$, whose second region is much smaller (Fig.~\ref{fig:redundancy_heatmap}).
Our pruning selection achieves the lowest degradation in nearly all evaluated settings, supporting approximate additivity as a practical basis for pruning selection.
Configurations selected on the calibration sets also perform well on held-out and unseen datasets, supporting their transferability and generalization.

\begin{figure}[b]
    \vspace{-1ex}
    \centering
    \includegraphics[width=0.99\linewidth]{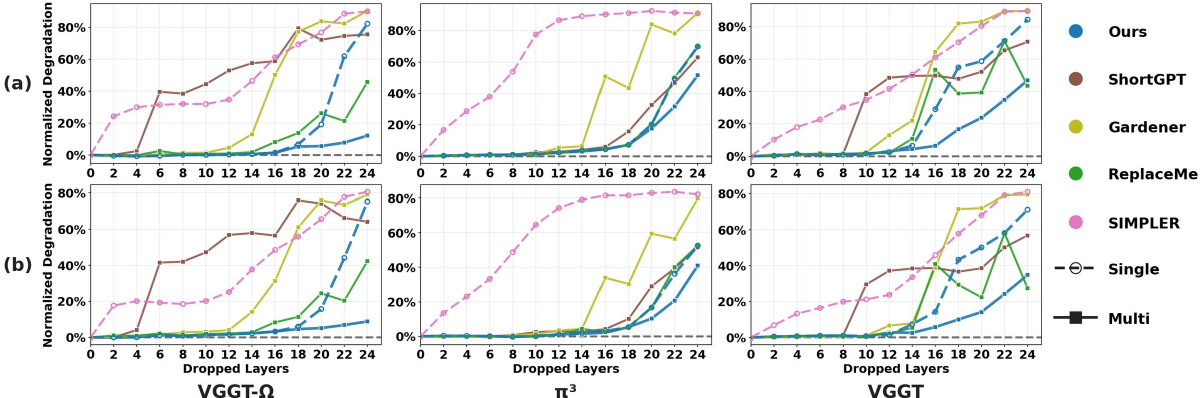}
    \vspace{-0.5em}
    \caption{\textbf{Pruning quality and generalization.}
    For each method and budget, one pruning configuration is selected based only on the combined 100-scene calibration set from ScanNet and nuScenes and applied unchanged to:
    (a) the held-out evaluation sets, and
    (b) the unseen 7Scenes and Waymo datasets.
    ``Single'' restricts removal to one contiguous interval, whereas ``Multi'' allows multiple intervals according to each method's search space; our method allows up to two separated intervals.
    }
    \label{fig:pruning_curves}
\end{figure}

\findingbox{2}{Degradation is approximately \textbf{additive} for deletions within the two redundancy regions, enabling effective two-interval pruning search with $O(L^2)$ model evaluations.}

\begin{figure}[t]
    \centering
    \includegraphics[width=0.99\linewidth]{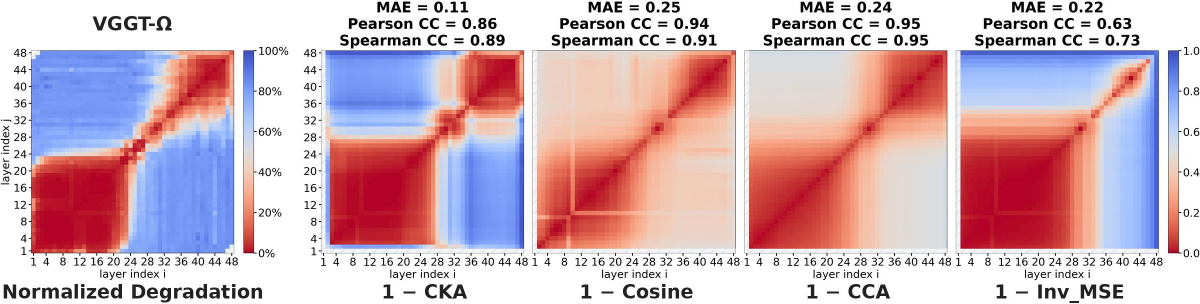}
    \vspace{-0.5em}
    \caption{\textbf{Representation-based proxies versus measured normalized degradation on VGGT-$\Omega$.}
    CKA achieves the lowest MAE, while cosine similarity and CCA yield higher correlations.
    }
    \label{fig:sim_heatmap_vggt_omega}
    \vspace{-1.5em}
\end{figure}

\subsection{One-Forward-Pass Representation Proxies}\vspace{-1ex}
\label{sec:cka}

\begin{figure}[b]
    \vspace{-1.2em}
    \centering
    \includegraphics[width=0.96\linewidth]{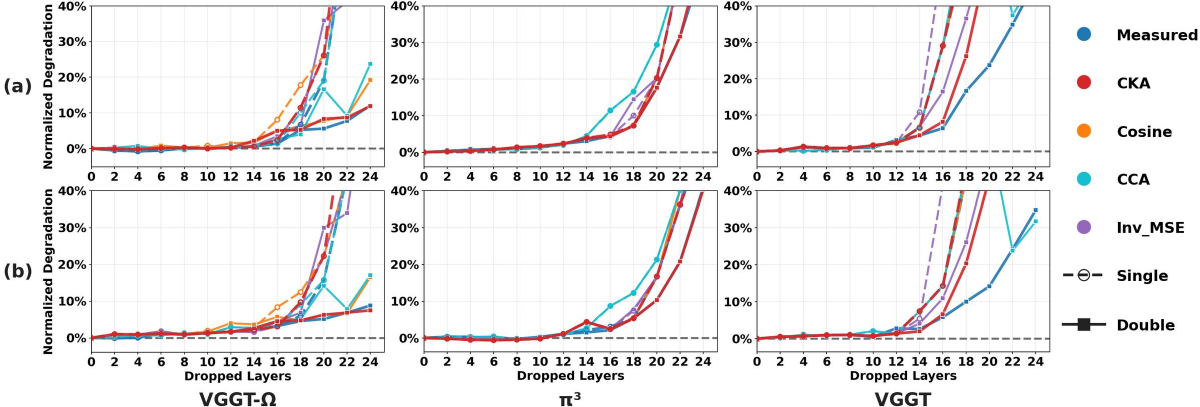}
    \vspace{-1em}
    \caption{\textbf{Pruning quality and generalization with representation-based proxies.}
    The protocol and row definitions follow Fig.~\ref{fig:pruning_curves}.
    }
    \label{fig:pruning_curves_sim}
\end{figure}

We further examine whether single-interval degradation admits a cheaper representation-based proxy.
Let $H_t$ denote the representations after layer $t$, with $H_0$ denoting the aggregator input.
For each interval $I=[i,j]$, we compare its input $H_{i-1}$ and output $H_j$ using CKA~\citep{cka}, cosine similarity, CCA, and inverse MSE.
Each measure $q$ is normalized to produce a similarity score $s_q(i,j)\in[0,1]$, giving the degradation proxy $\widetilde D_q(I)=1-s_q(i,j)$.

Fig.~\ref{fig:sim_heatmap_vggt_omega} compares these proxies with measured normalized degradation for VGGT-$\Omega$; results for the other two backbones appear in the appendix (Fig.~\ref{fig:sim_heatmap}).
All four broadly reflect the early and later redundancy regions, but differ in their boundaries and agreement with measured degradation.
Cosine similarity and CCA achieve Pearson and Spearman correlations above $0.90$, with MAEs around $0.25$.
CKA has slightly lower correlations around $0.9$, yet achieves the lowest MAE of $0.12$.
Inverse MSE shows weaker correlations and an MAE of $0.22$.

We then replace measured single-interval degradation with each proxy, keeping the pruning search space and evaluation protocol unchanged.
For two separated intervals $A=[i,j]$ and $B=[k,l]$, we use the additive estimate $\widetilde D_q(A\cup B)=\widetilde D_q(A)+\widetilde D_q(B)=2-s_q(i,j)-s_q(k,l)$.
Thus, representation-based selection approximates both single-interval degradation and the joint effect of two deletions.
Fig.~\ref{fig:pruning_curves_sim} shows that CKA is not always the best criterion for single-interval selection, but generally performs best among the four proxies when allowing up to two intervals.
It closely matches our measured-degradation criterion on VGGT-$\Omega$ and $\pi^3$, although a gap emerges on VGGT beyond 16 removed layers.
CKA's lower MAE on VGGT-$\Omega$ may help explain its effectiveness compared with cosine similarity and CCA under additive scoring, where numerical agreement with measured degradation matters beyond correlation alone.
Further details are provided in Appendix~\ref{app:Similarity}.

CKA itself requires only one intact-model forward pass per calibration input, followed by $O(L^2)$ low-cost similarity computations on cached features.
For a 48-layer aggregator and 100 calibration scenes, this reduces calibration from approximately 33 GPU hours under the preceding cost assumption to less than one hour, offering a practical trade-off between calibration cost and pruning quality.

\findingbox{3}{CKA provides a practical proxy for effective two-interval pruning, requiring only one intact-model forward pass per calibration input.}

\vspace{-0.3em}
\section{What Can Calibration Recover?}\vspace{-1ex}
\label{sec:recovery}
Removing an interval creates a representation mismatch at its output boundary.
A natural response is to use knowledge distillation~\citep{minitron} or parameter-efficient fine-tuning~\citep{llmpruner}.
% such as LoRA~\citep{lora} and QLoRA~\citep{qlora}; 
However, these approaches still require iterative training, with recovery quality depending on the available training data and compute.
We instead investigate whether lightweight mappings fitted on a small calibration set can recover accuracy without end-to-end retraining.
We identify a limitation of shared linear calibration in the VGGT family: heterogeneous token types may require different recovery mappings.
We theoretically characterize the resulting reconstruction penalty and develop an adaptive token-type-aware recovery method.
Experiments show that separate mappings with a shared fallback improve recovery on VGGT and VGGT-$\Omega$.

\subsection{Why Distinguish Token Types?}\vspace{-1ex}
\label{sec:role-theory}
VGGT and VGGT-$\Omega$ contain camera and register tokens serving global or auxiliary roles, alongside patch tokens representing dense visual features.
We refer to camera and register tokens collectively as \textit{special tokens}.
Patch tokens substantially outnumber special tokens and can therefore dominate a reconstruction objective that weights all tokens equally.
Beyond this imbalance, the two token types may require \textit{different} recovery mappings.
In that case, a shared mapping must compromise between them, even when their contributions to the objective are balanced.

We quantify this compromise by comparing the best shared linear mapping with the best separate mappings.
For a fixed interval and calibration context, let $r\in\{s,p\}$ index special and patch tokens, respectively, and let $x_r$ and $y_r$ denote input and target output row vectors.
Define the expected reconstruction loss, its optimal linear mapping, and the input second-moment matrix as
\begin{equation}
    L_r(W)=\mathbb E\lVert x_rW-y_r\rVert_2^2,
    \qquad
    T_r=\arg\min_W L_r(W),
    \qquad
    \Sigma_r=\mathbb E[x_r^\top x_r],
\end{equation}
where the expectation is over input--target pairs of token type $r$.
Assume finite second moments and $\Sigma_r\succ0$.
Let $\pi_s,\pi_p>0$ with $\pi_s+\pi_p=1$ weight the two token types, and set $Q_r=\pi_r\Sigma_r$ and $\Delta_T=T_s-T_p$.
The excess reconstruction loss of the best shared mapping over separate mappings is
\begin{equation}
\label{eq:role-conflict}
    \min_W\sum_{r\in\{s,p\}}\pi_rL_r(W)
    -\sum_{r\in\{s,p\}}\pi_rL_r(T_r)
    =
    \operatorname{tr}\!\left[
        \Delta_T^\top
        (Q_s^{-1}+Q_p^{-1})^{-1}
        \Delta_T
    \right].
\end{equation}
This penalty is strictly positive whenever $T_s\ne T_p$, even with balanced weights $\pi_s=\pi_p=1/2$.
Thus, balancing the contributions of the two token types does not eliminate the reconstruction penalty of a shared mapping.
This result concerns unregularized population reconstruction loss and motivates type-specific recovery.
The proof of \eqref{eq:role-conflict} is provided in Appendix~\ref{app:role-proof}.

\subsection{Adaptive Token-Type-Aware Recovery}
\label{sec:token_recovery}

For each interval $[i,j]$, let $X_{\mathrm{all}}=\operatorname{cat}(X_s,X_p)$ and $Y_{\mathrm{all}}=\operatorname{cat}(Y_s,Y_p)$ denote its input and output activation matrices, concatenated along the token dimension.
We fit mappings $W_g=\mathbf I+\Delta_g$ for $g\in\{\mathrm{all},s,p\}$, where $\mathbf I$ is the identity matrix.
Following the residual least-squares formulation of Ghost~\citep{ghostedlayers}, each fit minimizes
$\lVert X_gW_g-Y_g\rVert_F^2+\lambda\lVert W_g-\mathbf I\rVert_F^2$
by solving
\begin{equation}
\label{eq:recovery_fit}
    (X_g^\top X_g+\lambda \mathbf I)\Delta_g
    =X_g^\top(Y_g-X_g),
    \qquad \lambda=10^{-6}.
\end{equation}
The type-specific recovery applies
$Y'_{\mathrm{all}}=\operatorname{cat}(X_sW_s,X_pW_p)$
with tokens restored to their original positions.
Intervals are processed sequentially from shallow to deep.
For each interval, input activations are collected from the model with all preceding pruning and recovery operations applied, while target output activations are collected from the intact model.
% Each fit therefore accounts for preceding pruning and recovery operations.

For VGGT and VGGT-$\Omega$, special-token fits use only 5 and 17 tokens per frame, respectively, far fewer than patch-token fits.
In our experiments, the special-token Gram matrix $X_s^\top X_s$ sometimes exhibits a condition number on the order of $10^{16}$, indicating severe ill-conditioning in the unregularized fitting problem.
We therefore use the relative deviation
$\rho=\lVert W_s-W_{\mathrm{all}}\rVert_F/\lVert W_{\mathrm{all}}-\mathbf I\rVert_F$
as a heuristic safeguard.
This score measures the departure of the special-token mapping from the shared mapping relative to the magnitude of the shared correction.
We use the type-specific mappings when $\rho<20$ and otherwise apply $W_{\mathrm{all}}$ to all tokens.
Since $\pi^3$ has only patch tokens, it uses one mapping per interval without adaptive selection.

\begin{figure}[t]
    \centering
    \includegraphics[width=\linewidth]{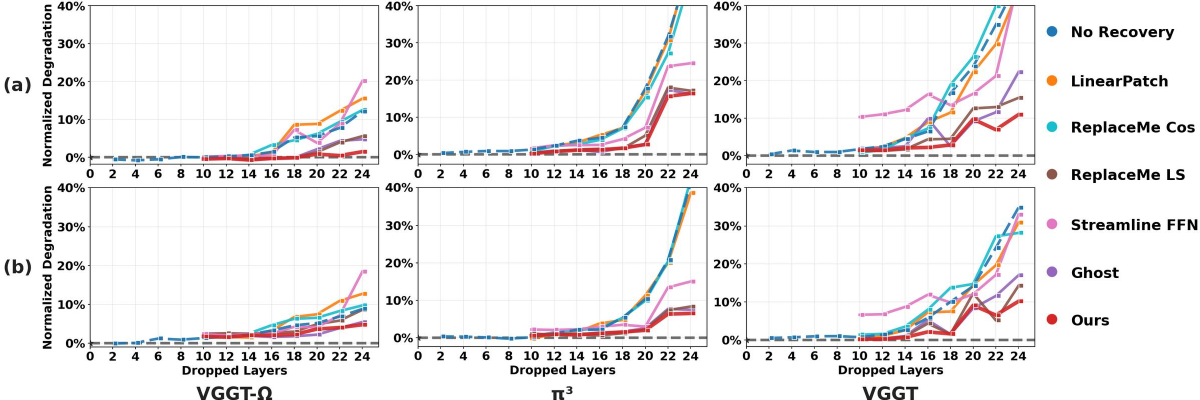}
    \vspace{-1.5em}
    \caption{\textbf{Post-pruning recovery and generalization.}
    Recovery is fitted on 100 ScanNet+nuScenes calibration scenes and evaluated on (a) held-out and (b) unseen 7Scenes+Waymo data.}
    % For each backbone and budget, all recovery methods use the same pruning configuration selected using measured degradation.
    \vspace{-1.5em}
    \label{fig:recovery_benchmark}
\end{figure}

Fig.~\ref{fig:recovery_benchmark} compares our method with LinearPatch~\citep{linearpatch}, ReplaceMe~\citep{replaceme}, Streamline~\citep{llmstreamline}, and Ghost~\citep{ghostedlayers}, using pruning configurations selected by our measured-degradation criterion and 100 calibration scenes from ScanNet and nuScenes.
For each recovery method, we report the better result between single- and two-interval pruning.
Implementation details are in Appendix~\ref{app:recovery}.
Streamline and ReplaceMe-Cosine often fail to improve over direct removal, whereas Ghost and ReplaceMe-LS substantially reduce degradation.
Our adaptive token-type-aware recovery further improves VGGT and VGGT-$\Omega$ in nearly all evaluated settings.
These gains persist on both held-out scenes and unseen datasets, showing that 100 calibration scenes suffice to fit recovery transformations that generalize beyond the calibration data.

\findingbox{4}{Linear calibration effectively recovers accuracy after pruning, with adaptive token-type-aware mappings providing further gains.}

\vspace{-0.3em}
\section{From Findings to Compressed Models}\vspace{-1ex}
\label{sec:final_results}

\textbf{Compression baselines.} We combine our findings to construct final compressed models and evaluate their accuracy and efficiency under a deployment-oriented protocol.
For each backbone, we use the calibration splits from all four datasets to select the pruning configuration and fit the recovery transformations, then freeze the model and evaluate it on their disjoint held-out splits.
We compare against ReplaceMe~\citep{replaceme}, which also combines layer removal with linear recovery, at the same retained depth.
We additionally compare and combine our method with FastVGGT~\citep{fastvggt} for token reduction and QuantVGGT~\citep{quantvggt} for quantization.
QuantVGGT is evaluated only on VGGT, without latency or memory results, because its released calibration parameters are backbone-specific and its public implementation does not provide INT4 deployment.

\begin{figure}[t]
\centering
\includegraphics[width=0.98\linewidth]{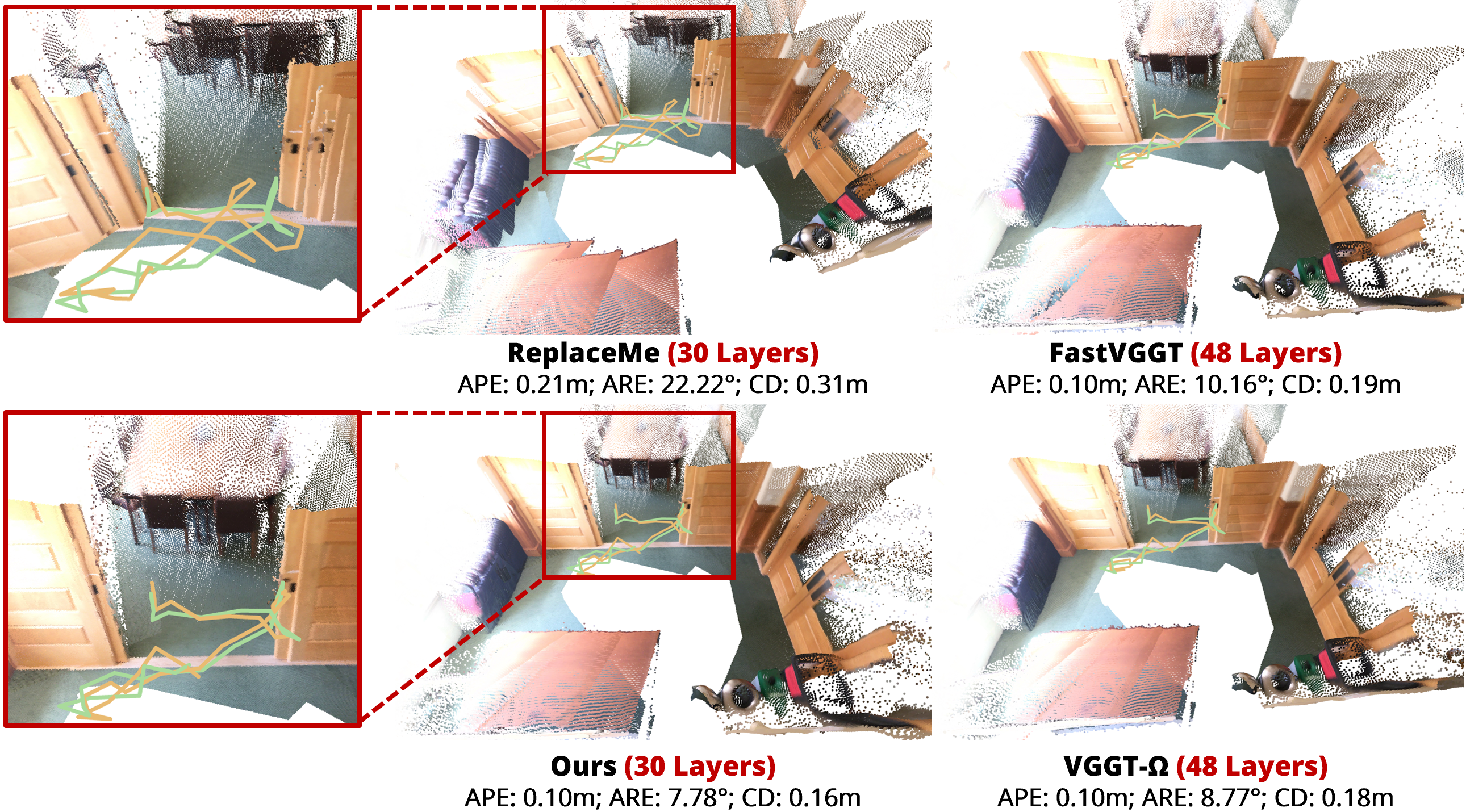}
\vspace{-0.5em}
\caption{\textbf{Qualitative reconstruction comparison.}
After removing 18 of 48 aggregator layers, ReplaceMe exhibits substantial geometric misalignment and trajectory errors,
while our recovery preserves accurate geometry and trajectories, matching or even improving upon the intact VGGT-$\Omega$.}
\label{fig:qualitative}
\end{figure}

\textbf{Accuracy and efficiency.} Table~\ref{tab:final_results} and Fig.~\ref{fig:qualitative} provide quantitative and qualitative comparisons, respectively.
For each backbone, we report a high pruning budget that preserves accuracy close to the intact model; at these budgets, measured-degradation and CKA-based selection choose the same pruning configuration.
Removing 18/48, 16/36, and 10/48 aggregator blocks from VGGT-$\Omega$, $\pi^3$, and VGGT reduces aggregator parameters by 20.8--44.5\%, latency by 14.2--30.9\%, and memory by 6.8--18.5\% on 30-frame inputs, where parameter counts refer to the aggregator only, while latency and peak allocated GPU memory are measured over the full forward pass.
In comparison, the token reduction and quantization methods considered here retain the original parameter count. These savings come with modest accuracy changes overall.
Averaged equally across the three backbones and the absolute and relative metrics within each category, translation and rotation errors increase by approximately 0.04\,m and 0.32$^\circ$, respectively, while CD increases by 0.01\,m.
Our method outperforms ReplaceMe across almost all metrics and backbones at matched retained depths.
Compared with FastVGGT and QuantVGGT, our method achieves the best accuracy in 18 of the 21 backbone--metric combinations.

The achievable compression is consistent with the deletion landscapes in Fig.~\ref{fig:redundancy_heatmap}.
VGGT-$\Omega$ and $\pi^3$ exhibit broad early redundancy regions and tolerate removing 18 and 16 blocks, respectively.
VGGT has a narrower early region and exhibits higher trajectory degradation even after removing only 10 blocks.
Adding our layer pruning to FastVGGT further reduces latency while maintaining broadly comparable accuracy to FastVGGT alone.
This supports their complementary effects on model depth and token computation, allowing both techniques to be used together.

\begin{table*}[t]
\centering
\vspace{-2.5ex}
\caption{\textbf{Comparison averaged over the held-out splits of ScanNet, 7Scenes, nuScenes, and Waymo.}
Bold error values indicate the best results among compressed variants of each backbone.
% excluding the intact model.
% Green subscripts indicate percentage reductions relative to the corresponding intact backbone, computed before rounding.
}
\label{tab:final_results}
\small
\setlength{\tabcolsep}{4.5pt}
\renewcommand{\arraystretch}{1.15}
\resizebox{\textwidth}{!}{%
\begin{tabular}{@{}lcccccccclll@{}}
\toprule
& & \multicolumn{4}{c}{Trajectory Errors}
& \multicolumn{3}{c}{Geometric Errors}
& \multicolumn{3}{c}{System Efficiency} \\
\cmidrule(lr){3-6}
\cmidrule(lr){7-9}
\cmidrule(lr){10-12}
Method
& Layers
& APE $\downarrow$
& ARE $\downarrow$
& RPE\textsubscript{t} $\downarrow$
& RPE\textsubscript{r} $\downarrow$
& Acc. $\downarrow$
& Comp. $\downarrow$
& CD $\downarrow$
& Params $\downarrow$
& Latency $\downarrow$
& Memory $\downarrow$ \\
& & (m) & ($^\circ$) & (m) & ($^\circ$)
& (m) & (m) & (m) & (M) & (s) & (GiB) \\
\midrule

\rowcolor{fullgray}
\textbf{VGGT-$\Omega$}
& 48
& 0.31 & 9.99 & 0.18 & 0.93
& 1.15 & 1.08 & 1.11
& 605 & 1.98 & 8.2 \\
$+$ FastVGGT
& 48
& 0.57 & 11.44 & 0.33 & 1.04
& 1.15 & 1.10 & 1.13
& 605
& 1.70\reduction{14.0}
& 8.2 \\
$+$ ReplaceMe
& \textbf{30}
& 0.96 & 23.46 & 0.53 & 4.59
& 1.38 & 1.30 & 1.34
& 378\reduction{37.5}
& 1.57\reduction{20.9}
& 6.9\reduction{15.9} \\
\rowcolor{oursblue}
$+$ \textbf{Ours}
& \textbf{30}
& \textbf{0.33} & \textbf{9.07} & \textbf{0.19} & \textbf{1.00}
& \textbf{1.13} & \textbf{1.09} & \textbf{1.11}
& 378\reduction{37.5}
& 1.58\reduction{20.4}
& 6.9\reduction{15.9} \\
$+$ Ours $+$ FastVGGT
& \textbf{30}
& 0.64 & 11.24 & 0.38 & 1.15
& 1.14 & \textbf{1.09} & 1.12
& 378\reduction{37.5}
& 1.39\reduction{29.5}
& 6.9\reduction{15.9} \\

\midrule

\rowcolor{fullgray}
$\boldsymbol{\pi}^{\textbf{3}}$
& 36
& 0.36 & 10.59 & 0.21 & 0.96
& 1.25 & 1.12 & 1.18
& 454 & 1.16 & 6.5 \\
$+$ FastVGGT
& 36
& 0.51 & 12.58 & 0.35 & \textbf{1.05}
& 1.30 & \textbf{1.14} & 1.22
& 454
& 0.95\reduction{18.4}
& 6.5 \\
$+$ ReplaceMe
& \textbf{20}
& 0.73 & 12.99 & 0.35 & 1.32
& \textbf{1.27} & 1.17 & 1.22
& 252\reduction{44.5}
& 0.79\reduction{32.3}
& 5.3\reduction{18.5} \\
\rowcolor{oursblue}
$+$ \textbf{Ours}
& \textbf{20}
& \textbf{0.37} & \textbf{10.92} & \textbf{0.22} & 1.13
& \textbf{1.27} & \textbf{1.14} & \textbf{1.20}
& 252\reduction{44.5}
& 0.80\reduction{30.9}
& 5.3\reduction{18.5} \\
$+$ Ours $+$ FastVGGT
& \textbf{20}
& 0.66 & 12.05 & 0.46 & 1.20
& 1.33 & 1.15 & 1.24
& 252\reduction{44.5}
& 0.69\reduction{40.9}
& 5.3\reduction{18.5} \\

\midrule

\rowcolor{fullgray}
\textbf{VGGT}
& 48
& 0.90 & 9.90 & 0.52 & 1.24
& 1.13 & 1.12 & 1.12
& 605 & 1.78 & 8.8 \\
$+$ FastVGGT
& 48
& 1.14 & 12.06 & 0.56 & 1.46
& 1.13 & \textbf{1.12} & \textbf{1.12}
& 605
& 1.40\reduction{18.8}
& 8.8 \\
$+$ QuantVGGT
& 48
& 1.12 & 14.25 & 0.79 & 1.56
& 1.16 & \textbf{1.12} & 1.14
& 605
& -- & -- \\
$+$ ReplaceMe
& \textbf{38}
& 1.23 & 12.27 & 0.51 & \textbf{1.27}
& \textbf{1.12} & 1.14 & 1.13
& 479\reduction{20.8}
& 1.44\reduction{19.2}
& 8.2\reduction{6.8} \\
\rowcolor{oursblue}
$+$ \textbf{Ours}
& \textbf{38}
& 1.12 & \textbf{12.05} & \textbf{0.48} & 1.38
& \textbf{1.12} & 1.15 & 1.13
& 479\reduction{20.8}
& 1.49\reduction{14.2}
& 8.2\reduction{6.8} \\
$+$ Ours $+$ FastVGGT
& \textbf{38}
& \textbf{1.06} & 13.75 & 0.58 & 1.58
& 1.14 & \textbf{1.12} & 1.13
& 479\reduction{20.8}
& 1.23\reduction{28.9}
& 8.2\reduction{6.8} \\
$+$ Ours $+$ QuantVGGT
& \textbf{38}
& 1.34 & 16.04 & 0.68 & 1.57
& 1.15 & 1.17 & 1.16
& 479\reduction{20.8}
& -- & -- \\

\bottomrule
\end{tabular}%
}
\vspace{-0.7em}
\end{table*}

\vspace{-0.7em}
\section{Conclusion and Discussion}
\vspace{-0.3em}
\label{sec:conclusion}
We identify two separated redundancy regions across the VGGT family, with a dominant early region and a narrower late one.
This structure enables effective pruning: approximate additivity reduces the required model evaluations from $O(L^4)$ to $O(L^2)$, while a CKA proxy requires only one intact-model forward pass per calibration input.
Together with adaptive token-type-aware recovery, the resulting compressed models substantially reduce parameter counts and inference latency while maintaining accuracy comparable to their intact counterparts across datasets and sequence lengths.
The origin of this shared redundancy pattern remains unclear.
Future work should distinguish among possible causes, including shared architecture design, multi-view geometric objectives, pretrained visual features, and optimization dynamics, ideally by tracking redundancy throughout training.
Such understanding could guide adaptive depth allocation or the training of shallower models, extending redundancy analysis beyond post-training compression to reduce both training and inference costs.
\newpage

\bibliography{iclr2027_conference}
\bibliographystyle{iclr2027_conference}

\newpage
\appendix

\section{Related Work}
\label{sec:related}

\subsection{3D Visual Geometry Transformers and Efficient Inference}
DUSt3R \citep{DUSt3R} pioneered feed-forward 3D reconstruction by directly predicting dense pointmaps from image pairs, with subsequent works \citep{mast3r,fast3r,must3r,mvdust3r,monst3r} extending this paradigm to larger sets or dynamic scenes.
VGGT \citep{vggt} further advances this line of work by predicting camera parameters and dense geometry within a unified framework.
Building on VGGT, $\pi^3$ \citep{pi3} removes the dependence on a reference view, while VGGT-$\Omega$ \citep{vggtomega} demonstrates predictable scaling with model capacity and data size.
Their computational and memory requirements have motivated a range of efficiency methods.
FastVGGT \citep{fastvggt}, Co-Me \citep{come}, HTTM \citep{httm}, and LiteVGGT \citep{litevggt} reduce computation through token pruning or merging.
AVGGT and RegimeVGGT \citep{avggt,regimevggt} exploit depth-dependent attention redundancy to reduce computation.
QuantVGGT \citep{quantvggt} and QVGGT \citep{qvggt} instead reduce numerical precision through post-training quantization.
These approaches largely preserve parameter count; our work complements them by removing entire layers, reducing both parameter count and computation for inference and subsequent fine-tuning.

\subsection{Layer Redundancy in Transformers}

LLM studies often report greater redundancy in \textit{middle-to-late} layers \citep{shortgpt,uidl,curseofdepth,whatmatters} across Llama \citep{llama2}, Qwen \citep{qwen}, Mistral \citep{mistral}, Phi-2 \citep{phi2}, and DeepSeek \citep{deepseek}.
More recently, \citet{rethinkingredundancy} show that identified redundancy patterns can change substantially with the calibration objective in Llama and Qwen models, challenging the view of redundancy as an intrinsic property of a pretrained network.
The ViT literature presents a more varied picture of cross-layer similarity.
DeepViT \citep{deepvit} and SIMPLER \citep{simpler} report deeper-layer feature stabilization in supervised ViTs and in self-supervised ViT-MAE \citep{mae} models, respectively.
However, broader analyses show that similarity patterns vary along several dimensions: architecture, across ViT \citep{vit}, PiT \citep{pit}, and Swin \citep{swin}, as analyzed by \citet{howvitworks}; training strategy, across DeiT \citep{deit}, MoCo \citep{mocov3}, and SimMIM \citep{simmim} on a common ViT-B backbone, as examined by \citet{darksecrets}; and model scale, within the MAE and BEiT \citep{beit} families, as studied by \citet{teachingmatters}.
Yet representational similarity alone does not establish functional redundancy, and whether these patterns extend to 3DVGTs remains unexplored.

\subsection{Layer Pruning and Recovery in Transformers}
Cosine-based representation similarity commonly guides layer pruning \citep{shortgpt,uidl,llmstreamline,minitron}.
LaCo \citep{laco} and ReplaceMe \citep{replaceme} compare cosine-based criteria with alternatives including CKA \citep{cka}, KL divergence, $L_2$ distance, and CCA, favoring cosine in their respective settings.
SIMPLER \citep{simpler} uses CKA to locate representational stabilization and prunes subsequent layers before downstream adaptation.
Beyond activation similarity, Gardener \citep{gardener} uses pretrained-weight entropy for data-free one-shot pruning.
However, more recent work \citep{rethinkingrelevance} reports weak correlations between cosine similarity and pruning-induced degradation, while \citet{rethinkingredundancy} find that different search strategies often converge to similar solutions under a fixed calibration objective.
Removing transformer layers can disrupt downstream representations and degrade performance.
% , yet methods such as ShortGPT \citep{shortgpt} and SLEB \citep{sleb} omit explicit recovery.
Natural solutions use additional training: Minitron \citep{minitron} employs knowledge distillation, while LLM-Pruner \citep{llmpruner} and UIDL \citep{uidl} use LoRA \citep{lora} and QLoRA \citep{qlora}, respectively.
Another line of work performs local recovery using calibration activations from the intact model, avoiding end-to-end retraining under the original task objective.
LLM-Streamline \citep{llmstreamline} trains a small replacement module to approximate the removed layers, while LinearPatch \citep{linearpatch} corrects activation-magnitude mismatch through Hadamard transformations and channel-wise scaling.
ReplaceMe \citep{replaceme} absorbs a linear transformation into the preceding block's FFN output projection, whereas Ghost \citep{ghostedlayers} directly maps the full boundary hidden state to the activation produced after each removed layer.

\begin{figure}[t]
    \centering
    \includegraphics[width=0.99\linewidth]{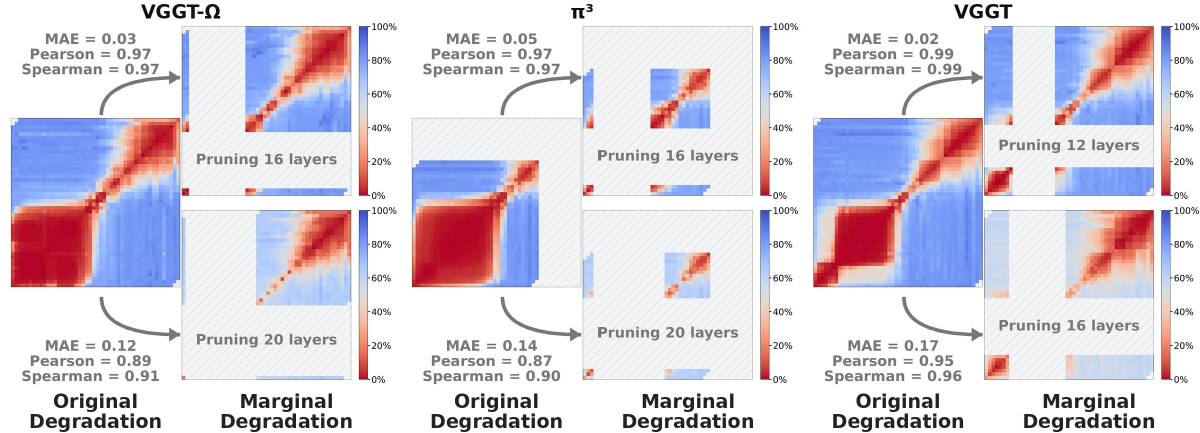}
    \caption{\textbf{Scope of approximate additivity.}
    Original degradation $D_{\mathcal M}(B)$ is compared with marginal degradation $D_{\mathcal M}(B\mid A)$ using the same normalization constants.
    Upper: shorter deletions within the early redundancy region largely preserve the original landscape.
    Lower: longer deletions reaching its boundary weaken the later redundancy region and reduce agreement.
    }
    \label{fig:marginal_heatmap_appendix}
\end{figure}

\begin{figure}[b]
    \centering
    \includegraphics[width=0.85\linewidth]{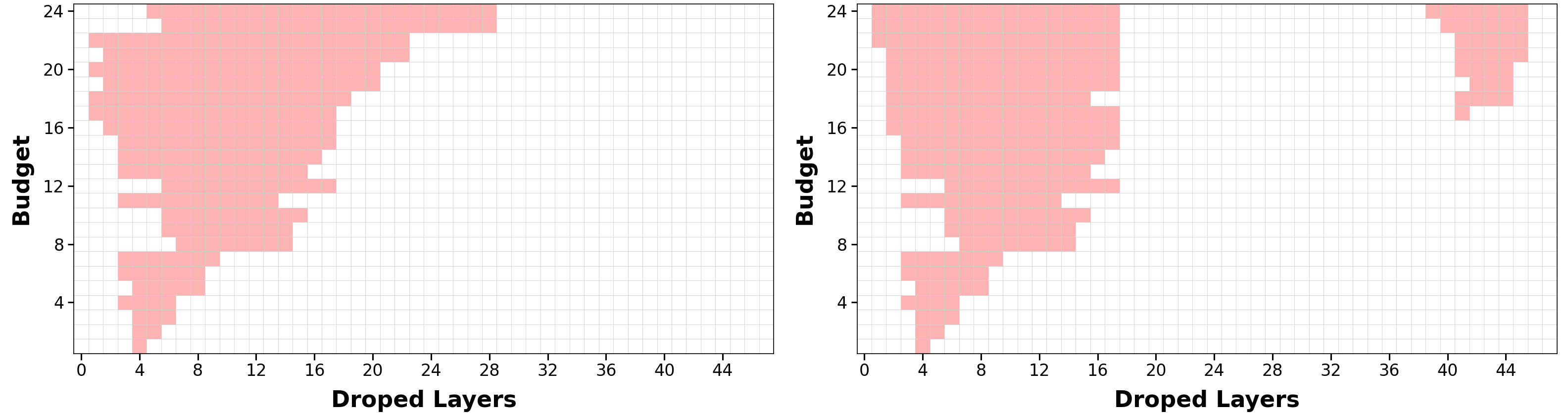}
    \caption{\textbf{Selected pruning intervals across budgets on VGGT-$\Omega$.}
Left: single-interval pruning progressively extends the early deletion as the budget grows.
Right: guided by approximate additivity, our two-interval selection keeps the early interval relatively stable and allocates additional removals to the later redundancy region, avoiding excessive pruning near the early-region boundary.}
    \label{fig:pruning_intervals}
\end{figure}

\section{Scope of Approximate Additivity}
\label{app:additivity}

Section~\ref{sec:composition} introduces the additive approximation and demonstrates its effectiveness for pruning selection.
Here, we further investigate when it remains accurate and when it begins to break down.
For each model, we select two early intervals $A$: a shorter interval lying well within the dominant redundancy region and a longer interval reaching its boundary.
For each choice, we evaluate the marginal degradation $D_{\mathcal M}(B\mid A)$ for intervals $B$ separated from $A$, using the calibration sets of all four datasets and the original normalization constants.

Fig.~\ref{fig:marginal_heatmap_appendix} compares the original and marginal degradation landscapes.
For shorter deletions lying within the dominant region (upper row), the later low-degradation region remains largely unchanged, with MAE of $0.02$--$0.05$ and Pearson and Spearman correlations of at least $0.97$.
For longer deletions reaching the early region's boundary (lower row), parts of the later region become less redundant.
MAE increases to $0.12$--$0.17$, while Pearson and Spearman correlations fall to as low as $0.87$ and $0.90$, respectively.
These results indicate that sufficiently extensive early deletions alter the effect of subsequent removals, weakening the additive approximation.

These results suggest avoiding deletions that exhaust the dominant early redundancy region when using the additive approximation.
Such extensive removals are usually unnecessary at moderate compression levels.
At larger budgets, two-interval pruning can instead allocate additional removals to the later redundancy region.
Fig.~\ref{fig:pruning_intervals} illustrates this behavior: single-interval pruning must progressively extend the early deletion as the budget grows, whereas two-interval pruning keeps the early interval relatively stable and places additional removals in the later region.
This both avoids the boundary regime where additivity becomes less accurate and helps explain the advantage of two-interval pruning at larger budgets.

\section{Pruning Implementation Details}
\label{app:Pruning}

\paragraph{Common protocol.}
We replace selected aggregator layers by identity operations and evaluate the resulting network.
Every method uses the same ordered layer catalog and is capped at 24 removed layers.
Reported curves evaluate budgets
$b\in\{2,4,\ldots,24\}$.
The similarity criteria within our framework use the token-region selection described in Appendix~\ref{app:token_region}.
External baselines retain their original scoring and selection designs.
For methods originally defined on LLM or ViT layers, we map their scoring and ranking rules to the individual aggregator layers of each backbone.
Other adaptations are stated below.

\paragraph{Measured-degradation and similarity-based selectors.}
The measured-degradation selector evaluates all single-interval removals and computes their normalized degradation $D_{\mathcal{M}}$.
The single-interval variant selects the interval with minimum normalized degradation at each budget.

The CKA, cosine, inverse-MSE, and CCA selectors use the same search procedure, replacing measured normalized degradation with the corresponding degradation proxy
$\widetilde{D}_q([i,j])=1-s_q(i,j)$.
For selection allowing up to two intervals, a candidate contains either one interval or two non-overlapping intervals separated by at least one retained layer.
The two-interval score is the sum of the constituent single-interval scores.
For similarity-based selection, this gives
\begin{equation}
    \widetilde{D}_q
    \bigl([i,j]\cup[k,l]\bigr)
    =
    2-s_q(i,j)-s_q(k,l).
\end{equation}

\paragraph{ShortGPT \cite{shortgpt}.}
Following the original paper and official implementation, we use one-shot block-influence ranking.
For layer $\ell$, we rank its influence using
\begin{equation}
    \mathrm{BI}_\ell
    =
    1-S_{\mathrm{cos}}(H_{\ell-1},H_\ell).
\end{equation}
Because $S_{\mathrm{cos}}$ is remapped to $[0,1]$, this score is a positive rescaling of the influence defined using unremapped cosine similarity and preserves the ranking.
Layers are sorted once by ascending influence, equivalently descending input--output cosine similarity, with the lower layer index breaking ties.
A budget-$b$ configuration removes the first $b$ layers in this ordering.
ShortGPT does not constrain the number or location of the resulting intervals and does not recompute influence after a layer has been removed.

\paragraph{Data-free Gardener \cite{gardener}.}
For every aggregator layer, we concatenate all recursively named rank-two parameters, covering the attention projections and the two MLP linear layers.
Following the released implementation, we compute exact-value weight-number entropy rather than histogram-bin entropy:
\begin{equation}
    E_\ell
    =
    -\sum_v p_{\ell,v}\log p_{\ell,v},
\end{equation}
where $p_{\ell,v}$ is the frequency of the exact floating-point value $v$ in layer $\ell$.
Layers are ranked once by ascending entropy, ties are resolved by layer index, and each pruning configuration is a prefix of this ranking.
Entropy is not recomputed after pruning.

\paragraph{ReplaceMe \cite{replaceme}.}
We reproduce the released angular-distance scan.
For an interval of length $r$ starting at layer $i$, its score is
$d_{\mathrm{ang}}(H_{i-1},H_{i+r-1})$.
Candidates are generated in start-index order, stably sorted by increasing distance, and greedily accepted if they do not overlap a previously selected interval.
The single variant selects one interval.
The double variant selects two equal-length intervals and therefore supports only even total pruning budgets.
Unlike the separated-interval search in our framework, this baseline allows adjacent intervals, whose union forms a single contiguous removal.
The mean of the selected distances is recorded for diagnostics but is not used as a joint optimization objective.
Distances are not recomputed after the first interval is selected.

\paragraph{SIMPLER \cite{simpler}.}
Let $\mathbf{S}$ denote its CKA similarity matrix over layer outputs.
For a cutoff retaining layers $1,\ldots,c$, define the retained and pruned submatrices as
\begin{equation}
    \mathbf{S}_{\mathrm{TL}}
    =\mathbf{S}_{1:c,\,1:c},
    \qquad
    \mathbf{S}_{\mathrm{BR}}
    =\mathbf{S}_{c+1:L,\,c+1:L},
\end{equation}
where the index ranges are inclusive.
We compute the mean absolute difference between consecutive rows,
$\delta(\cdot)$, and record the cutoff score
\begin{equation}
    g(c)
    =
    \delta(\mathbf{S}_{\mathrm{TL}})
    -
    \delta(\mathbf{S}_{\mathrm{BR}}).
\end{equation}
For a fixed budget $b$, suffix pruning uniquely determines the removed interval as $[L-b+1,L]$.
We therefore report the full suffix-pruning curve rather than selecting a different-shaped interval using the cutoff score.

\section{Similarity Implementation and Visualization}
\label{app:Similarity}

\paragraph{Activation collection.}
Following the main text, we index the $L$ aggregator layers from $1$ to $L$ in execution order, treating each complete frame or global transformer block as a separate pruning unit.
Let $A_0$ denote the aggregator input and $A_\ell$ the output of layer $\ell$.
For a scene,
$A_\ell\in\mathbb{R}^{B\times S\times P\times C}$,
where $B$ is the batch size, $S$ the number of frames, $P$ the number of tokens per frame, and $C$ the channel dimension.
We run the intact model once per calibration scene and retain the aggregator input and the output of every pruning unit.
Similarities are computed on the GPU in FP32 after flattening the batch, frame, and token axes:
\begin{equation}
    H_\ell=\operatorname{flat}(A_\ell)
    \in\mathbb{R}^{N\times C},
    \qquad N=BSP.
\end{equation}
Thus, $H_0$ denotes the flattened aggregator input, consistently with the main text.
No token subsampling is used in the reported runs.

\paragraph{Token regions and aggregation.}
\label{app:token_region}
For VGGT and VGGT-$\Omega$, we compute similarity matrices over three token regions: special tokens (camera and register tokens), patch tokens, and all tokens jointly.
Matrices are computed independently per scene and then averaged equally across scenes within each dataset and across calibration datasets.
For each similarity measure $q$, we consider the all-token matrix
$\mathbf{S}_q^{\mathrm{all}}$ and the equally weighted special--patch matrix
\begin{equation}
    \mathbf{S}_q^{\mathrm{s-p}}
    = \frac{1}{2}
    \left(
        \mathbf{S}_q^{\mathrm{special}}
        + \mathbf{S}_q^{\mathrm{patch}}
    \right).
\end{equation}
For each measure, pruning variant, and budget, we derive one pruning configuration from each matrix.
We evaluate both configurations on the calibration data and retain the one with lower measured normalized degradation, averaged over all seven metrics and the calibration datasets.
The selected configuration is frozen for subsequent evaluation.

This selection applies to CKA, cosine similarity, CCA, and inverse MSE within our framework.
External pruning baselines retain their original scoring and selection procedures.
Since $\pi^3$ contains only patch tokens, it bypasses token-region splitting, averaging, and selection.
Computing the similarity proxies requires one intact-model forward pass per calibration input.
Token-region selection additionally evaluates the two resulting pruned models on the calibration data.

\paragraph{Similarity measures.}
For two flattened activation matrices
$X,Y\in\mathbb{R}^{N\times C}$,
we use the following implementations.
Cosine similarity is the mean token-wise cosine, remapped from $[-1,1]$ to $[0,1]$:
\begin{equation}
    S_{\mathrm{cos}}(X,Y)
    =
    \frac{1}{2}
    \left(
        1+\frac{1}{N}
        \sum_{n=1}^{N}
        \frac{x_n^\top y_n}
        {(\lVert x_n\rVert_2+\epsilon)
         (\lVert y_n\rVert_2+\epsilon)}
    \right),
    \qquad \epsilon=10^{-8}.
\end{equation}
The angular distance used by ReplaceMe is
\begin{equation}
    d_{\mathrm{ang}}(X,Y)
    =
    \frac{1}{N}
    \sum_{n=1}^{N}
    \frac{
        \arccos\!\left(
            \operatorname{clip}(\cos(x_n,y_n),-1,1)
        \right)
    }{\pi}.
\end{equation}

For linear CKA, we center each feature dimension,
$\bar X=X-\mathbf{1}\mu_X^\top$ and
$\bar Y=Y-\mathbf{1}\mu_Y^\top$,
and compute
\begin{equation}
    S_{\mathrm{CKA}}(X,Y)
    =
    \frac{\lVert\bar X^\top\bar Y\rVert_F^2}
    {\max\!\left(
        \sqrt{
            \lVert\bar X^\top\bar X\rVert_F^2
            \lVert\bar Y^\top\bar Y\rVert_F^2
        },
        \epsilon
    \right)}.
\end{equation}
Inverse MSE is
\begin{equation}
    S_{\mathrm{invMSE}}(X,Y)
    =
    \left(
        1+\frac{\lVert X-Y\rVert_F^2}{NC}
    \right)^{-1}.
\end{equation}
Our CCA score is an SVCCA-style implementation:
each centered activation is first projected onto at most 64 principal components, covariance matrices are regularized by $10^{-5}\mathbf I$, and the score is the mean canonical correlation, clipped to $[0,1]$.

\paragraph{From activation similarity to an interval score.}
The uppercase notation $S_q(X,Y)$ denotes a similarity measure evaluated on two activation matrices, whereas $s_q(i,j)$ denotes the similarity score associated with pruning interval $[i,j]$.
All interval indices are inclusive.
Deleting $[i,j]$ bypasses layers $i,\ldots,j$, whose input and output representations are $H_{i-1}$ and $H_j$, respectively.
We define
\begin{equation}
    s_q(i,j)
    =
    S_q\!\left(H_{i-1},H_j\right),
    \qquad 1\leq i\leq j\leq L,
\end{equation}
where
$q\in\{\mathrm{cos},\mathrm{CKA},\mathrm{invMSE},\mathrm{CCA}\}$.
As in the main text, the corresponding degradation proxy is
\begin{equation}
    \widetilde{D}_q([i,j])
    =1-s_q(i,j).
\end{equation}
Pairwise matrices are saved per scene and after aggregation, allowing the similarity-based methods to reuse the same cached activations and ordered layer catalog.

\begin{figure}[t]
    \centering
    \includegraphics[width=0.98\linewidth]{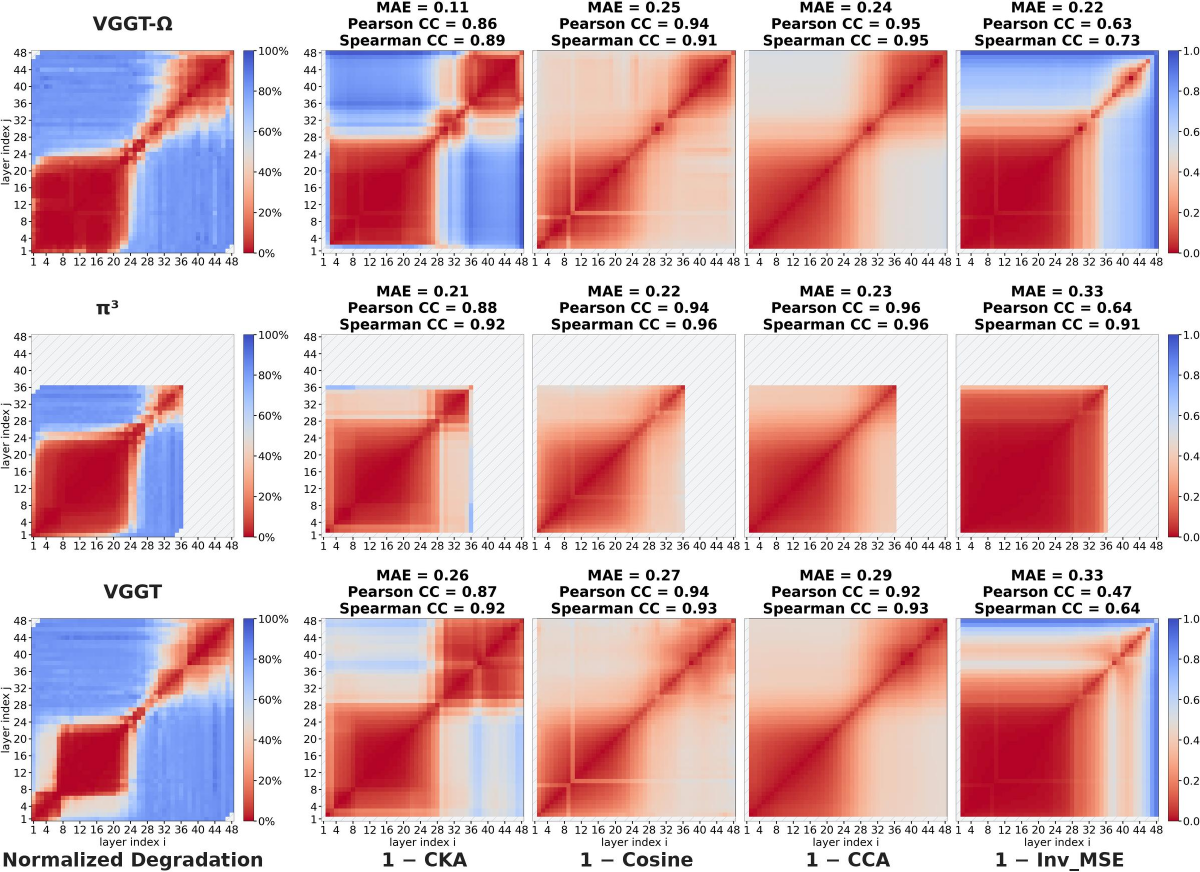}
    \vspace{-0.5em}
    \caption{\textbf{Representation similarity versus measured normalized degradation.}}
    \label{fig:sim_heatmap}
    \vspace{-0.7em}
\end{figure}

\paragraph{Similarity visualization.}
Fig.~\ref{fig:sim_heatmap} compares these proxies $\widetilde{D}_q$ with directly measured normalized degradation ${D}$.
All four measures capture the broad early and narrow later redundancy regions, although they differ from ${D}$ in region boundaries, numerical values, trends, and rankings.
CKA, cosine similarity, and CCA show strong agreement with measured degradation, while their relative performance varies across backbones and criteria.
These results suggest representation similarity as a proxy for normalized degradation.

\section{Recovery Implementation Details}
\label{app:recovery}

\paragraph{Baseline provenance and adaptation.}
The recovery baselines are ported from their original papers and released implementations.
We preserve their objectives, operator parameterizations, and default hyperparameters unless an adaptation is explicitly described below.
The batch, frame, and token axes of the collected activations are flattened into sample rows.
Activation sources, fitting order, and deployment differ across methods and are specified individually below.
Closed-form normal-equation statistics are accumulated in FP64.
Fitted linear operators are stored and applied in FP32, and their outputs are cast back to the residual-stream dtype.

\paragraph{LinearPatch \cite{linearpatch}.}
Following the original formulation and its released reference implementation, LinearPatch uses a normalized Sylvester Hadamard basis $Q$.
Our width-compatibility adaptation for non-power-of-two channel dimensions constructs the next larger Hadamard matrix, crops its leading $C\times C$ submatrix, and orthogonalizes it by QR decomposition.

For a single contiguous removal $[i,j]$, we collect
$X=H_{i-1}$ and $Y=H_j$ from the intact model and fit one operator across the interval.
For removals comprising multiple contiguous components, the implementation used in our experiments fits one operator per removed layer $\ell$, using
$X=H_{\ell-1}$ and $Y=H_\ell$ from the intact model.
In each case, $X,Y\in\mathbb{R}^{N\times C}$.
Channel scales and the resulting transform are
\begin{equation}
    s_c
    =
    \frac{\sum_n |(YQ)_{nc}|}
         {\sum_n |(XQ)_{nc}|+10^{-8}},
    \qquad
    W=Q\,\mathrm{diag}(s)Q^\top.
\end{equation}
Inference applies $XW$ at the corresponding replacement location.

\paragraph{Ghost \cite{ghostedlayers}.}
Following the official implementation, the baseline fits one matrix independently for each removed layer $\ell$.
Its source and target activations are
$X=H_{\ell-1}$ and $Y=H_\ell$, collected from the intact model.
It fits an identity-centered ridge regression to the boundary residual:
\begin{equation}
    (X^\top X+\lambda \mathbf I)\Delta
    =
    X^\top(Y-X),
    \qquad
    W=\mathbf I+\Delta,
\end{equation}
with $\lambda=10^{-6}$.
Each removed layer is replaced by its fitted linear transformation.
All matrices are fitted independently from the intact-model activation cache and applied in model order.

\paragraph{ReplaceMe-LS and ReplaceMe-Cosine \cite{replaceme}}
ReplaceMe modifies the MLP residual of the source layer rather than adding an external boundary module.
For a removed interval $[i,j]$, the source is layer $i-1$ and the target is the output of layer $j$.
Let $H$ be the source layer output, $R$ its MLP residual, and $Y$ the target output.
Replacing $R$ by $RA$ gives $H-R+RA$, so the desired transformed residual is
$T=Y+R-H$.
ReplaceMe-LS solves
\begin{equation}
    (R^\top R+\alpha \mathbf I)A
    =
    R^\top T,
\end{equation}
with the default $\alpha=0$.
Each interval is fitted independently using activations from the intact model.

ReplaceMe-Cosine initializes $A=\mathbf I$ and minimizes
$1-\operatorname{mean}_n\cos((RA)_n,T_n)$
using Adam for 10 epochs, a row batch size of 1,024, learning rate $10^{-4}$, and seed 0.
For multiple intervals, it follows the released front-to-back pipeline:
$H$, $R$, and $Y$ for a later interval are collected from the current model after preceding replacements and pruning have been installed.

For both variants, the fitted transform is fused into the source MLP's second linear layer, including the appropriate LayerScale change of basis when present, and therefore introduces no runtime hook.
The choices $\alpha=0$, 10 epochs, row batch size 1,024, and learning rate $10^{-4}$ are retained from the released implementation; the seed is fixed for reproducibility.

\paragraph{LLM-Streamline FFN \cite{llmstreamline}.}
We retain the released replacement-network architecture and optimization settings while adapting its training target to an activation-boundary objective for the VGGT-family backbones.
For each maximal contiguous removed interval $[i,j]$, we collect
$X=H_{i-1}$ and $Y=H_j$ from the intact model.
Each interval receives a replacement network
\begin{equation}
    f(X)
    =
    \operatorname{ReLU}(XW_1+b_1)W_2+b_2,
    \qquad C\rightarrow4C\rightarrow C,
\end{equation}
where $W_1\in\mathbb{R}^{C\times4C}$,
$W_2\in\mathbb{R}^{4C\times C}$,
and biases are broadcast across sample rows.
The network is trained to minimize
$\operatorname{MSE}(f(X),Y)$.

We use AdamW for one epoch with row batch size 16,384, learning rate $2\times10^{-4}$, minimum learning rate $5\times10^{-6}$, weight decay $10^{-3}$, betas $(0.9,0.95)$, seed 0, and a 3\% linear warmup followed by cosine decay.
Replacement networks are fitted independently from the intact-model activation cache and applied in model order at inference.

\section{Proof of the Shared-Mapping Reconstruction Penalty}
\label{app:role-proof}

We prove ~\eqref{eq:role-conflict} for a fixed interval and calibration context.
For each token type $r\in\{s,p\}$, let
$L_r(W)=\mathbb E\lVert x_rW-y_r\rVert_2^2$,
$\Sigma_r=\mathbb E[x_r^\top x_r]\succ0$, and
$T_r=\arg\min_W L_r(W)$.
We assume finite second moments and positive weights $\pi_s,\pi_p$ summing to one.
Define $Q_r=\pi_r\Sigma_r$ and $\Delta_T=T_s-T_p$.

\paragraph{Quadratic decomposition.}
The least-squares optimum satisfies the normal equation
$\Sigma_rT_r=\mathbb E[x_r^\top y_r]$.
Expanding the loss around $T_r$ therefore eliminates the cross term and gives
\begin{equation}
    L_r(W)
    =
    L_r(T_r)
    +\operatorname{tr}\!\left[
        (W-T_r)^\top\Sigma_r(W-T_r)
    \right].
\end{equation}
Consequently, the excess loss of the optimal shared mapping over separate mappings is
\begin{equation}
    \mathcal B_{\mathrm{role}}
    =
    \min_W\sum_{r\in\{s,p\}}
    \operatorname{tr}\!\left[
        (W-T_r)^\top Q_r(W-T_r)
    \right].
\end{equation}

\paragraph{Optimal shared mapping.}
Let $S=Q_s+Q_p$.
Differentiating the quadratic objective yields
$W^*=S^{-1}(Q_sT_s+Q_pT_p)$.
Writing $Z=W-T_p$, the same objective becomes
\begin{align}
    &\operatorname{tr}\!\left[
        (Z-\Delta_T)^\top Q_s(Z-\Delta_T)
        +Z^\top Q_pZ
    \right] \nonumber\\
    &\quad=
    \operatorname{tr}\!\left[
        (Z-S^{-1}Q_s\Delta_T)^\top
        S(Z-S^{-1}Q_s\Delta_T)
    \right]
    +\operatorname{tr}\!\left[
        \Delta_T^\top(Q_s-Q_sS^{-1}Q_s)\Delta_T
    \right].
\end{align}
The first term vanishes at the optimum.
Moreover,
$Q_s-Q_sS^{-1}Q_s=Q_sS^{-1}Q_p$,
whose inverse is
$Q_p^{-1}SQ_s^{-1}=Q_p^{-1}+Q_s^{-1}$.
Thus,
\begin{equation}
    \mathcal B_{\mathrm{role}}
    =
    \operatorname{tr}\!\left[
        \Delta_T^\top
        (Q_s^{-1}+Q_p^{-1})^{-1}
        \Delta_T
    \right],
\end{equation}
which proves ~\eqref{eq:role-conflict}.
Since $(Q_s^{-1}+Q_p^{-1})^{-1}$ is positive definite, this penalty is zero if and only if $T_s=T_p$.

\paragraph{Token imbalance and balanced fitting.}
For the illustrative case $\Sigma_s=\Sigma_p=\Sigma$, the shared optimum reduces to
$W^*=\pi_sT_s+\pi_pT_p$.
Defining $D_T=\lVert\Sigma^{1/2}(T_s-T_p)\rVert_F^2$, we obtain
\begin{equation}
    \mathcal B_{\mathrm{role}}=\pi_s\pi_pD_T,
    \qquad
    L_s(W^*)-L_s(T_s)=\pi_p^2D_T,
    \qquad
    L_p(W^*)-L_p(T_p)=\pi_s^2D_T.
\end{equation}
When $\pi_s$ is small, the pooled excess loss can be small even though the special-token excess remains close to $D_T$.
Balanced weights remove this asymmetry but still incur a shared-mapping penalty of $D_T/4$.
Thus, reweighting alone need not resolve disagreement between the optimal mappings.

\paragraph{Scope of the analysis.}
The result concerns unregularized population reconstruction loss within the linear mapping class; the deleted computation itself need not be linear.
Our implementation instead fits identity-centered ridge mappings from finite calibration samples.
Separate mappings remove the shared-mapping constraint but may incur greater estimation error, particularly for the substantially less numerous special tokens.
The analysis therefore motivates token-specific fitting and an adaptive safeguard, but does not derive our selection criterion or threshold, nor guarantee improved downstream geometric accuracy.

\section{Calibration and runtime cost}
All experiments are conducted on a single NVIDIA RTX 6000 Ada GPU with 48\,GB memory.
The main calibration cost comes from pruning selection.
Without approximation, exhaustive two-interval search requires $O(L^4)$ model evaluations.
Our approximate additivity reduces this to evaluating only the $\binom{L+1}{2}=O(L^2)$ single intervals, i.e., 1,176 pruned configurations per scene for the 48-layer VGGT and VGGT-$\Omega$, and 666 for the 36-layer $\pi^3$.
For 100 calibration scenes, the measured wall-clock times are approximately 17 hours for VGGT, 11 hours for VGGT-$\Omega$, and 6 hours for $\pi^3$.
CKA-based selection further reduces the calibration cost.
Including cached-feature similarity computation and the additional evaluations used for token-region selection, calibration on 100 scenes takes less than one hour for each backbone.
These wall-clock times are rough measurements under our current implementation and hardware setup.
They depend on implementation details, such as the degree of overlap between GPU inference and CPU-side evaluation, as well as dataset characteristics such as image resolution and preprocessing cost, and therefore need not scale directly with the number of model evaluations.
Post-pruning recovery adds little additional cost: the recovery mappings are obtained by closed-form least squares without iterative training, with fitting taking only a few seconds.
At inference, each recovery transformation adds only a matrix multiplication at a pruning boundary, whose overhead is negligible relative to the retained transformer computation.

\section{Ablation Study}
\label{app:calibration}

\begin{figure}[t]
    \centering
    \includegraphics[width=\linewidth]{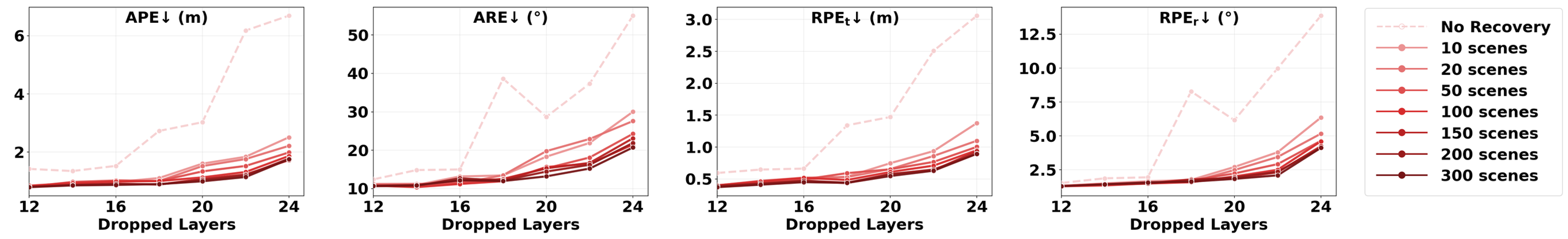}
    \caption{\textbf{Effect of calibration-set size on post-pruning recovery.}
    % Calibration scenes are drawn equally from ScanNet and nuScenes; evaluation uses their held-out splits and unseen 7Scenes and Waymo.
    }
    \label{fig:calibration_size}
\end{figure}

\paragraph{Calibration-set Size.}
We examine post-pruning recovery using equal numbers of ScanNet and nuScenes calibration scenes, evaluating on their held-out splits and the unseen 7Scenes and Waymo datasets.
Fig.~\ref{fig:calibration_size} shows that even 10 scenes substantially reduce pruning-induced errors.
Increasing the size from 10 to 100 scenes generally improves all seven metrics, especially at higher pruning budgets.
Beyond 100 scenes, trends across adjacent budgets become smoother, but accuracy gains are limited.
We therefore use 100 scenes for the recovery benchmark to balance cost and accuracy.

\begin{figure}[t]
    \centering
    \includegraphics[width=\linewidth]{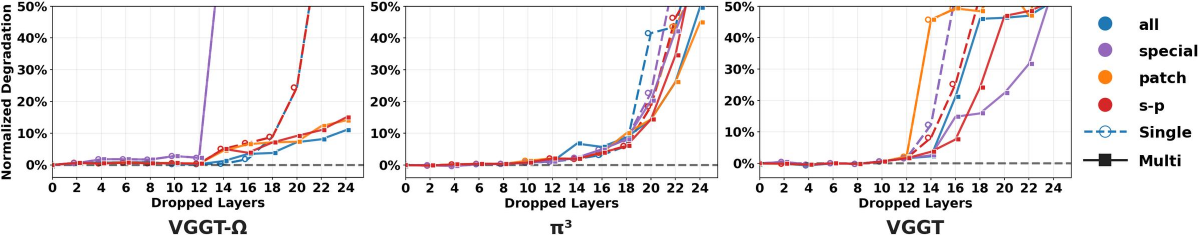}
    \caption{\textbf{Effect of token-region handling on similarity-based pruning.} 
    Using special or patch token type alone generally performs poorly. Neither the all-token nor the special--patch mean variant consistently dominates, motivating selection based on measured calibration degradation.
}
    \label{fig:token_ablation}
\end{figure}

\paragraph{Token-region Selection.}
\label{sec:token_region_ablation}
We compare four similarity-based pruning variants for VGGT and VGGT-$\Omega$: special tokens only, patch tokens only, all tokens jointly, and the mean of separately computed special-token and patch-token similarities.
Fig.~\ref{fig:token_ablation} shows that using either token type alone generally performs poorly, while neither the all-token nor the special--patch mean variant consistently outperforms the other.
Given the low computational cost of similarity-based scoring, we generate pruning configurations from both latter variants and select the one with lower measured degradation on the calibration data.
The selected configuration is then frozen for evaluation.
Implementation details are provided in Appendix~\ref{app:token_region}.

\section{Long-sequence Generalization}
\label{app:long}
We further evaluate whether pruning configurations and recovery transformations calibrated on 30-frame sequences generalize to longer inputs.
We use the same compressed models as in Table~\ref{tab:final_results}, calibrated only on 30-frame sequences, and apply them unchanged to longer inputs.
Table~\ref{tab:long_sequence} reports results on held-out scenes using 30, 100, and 200 frames.
Across all three backbones, the compressed models retain comparable accuracy while preserving parameter, latency, and memory savings, demonstrating generalization beyond the calibration sequence length.

\begin{table*}[t]
\centering\vspace{-4ex}
\caption{\textbf{Generalization across scenes and sequence lengths.}
Pruning configurations and recovery transformations are calibrated using only 30-frame sequences and applied unchanged to held-out test scenes with 30, 100, and 200 frames.
The results demonstrate generalization to both unseen scenes and longer sequences.
Parameter counts include only the aggregator, whereas latency and peak allocated GPU memory are measured over the complete forward pass.
}
\label{tab:long_sequence}
\small
\setlength{\tabcolsep}{4.5pt}
\renewcommand{\arraystretch}{1.15}
\resizebox{\textwidth}{!}{%
\begin{tabular}{@{}lccccccccclll@{}}
\toprule
& & & \multicolumn{4}{c}{Trajectory Errors}
& \multicolumn{3}{c}{Geometric Errors}
& \multicolumn{3}{c}{System Efficiency} \\
\cmidrule(lr){4-7}
\cmidrule(lr){8-10}
\cmidrule(lr){11-13}
Method
& Frames
& Layers
& APE $\downarrow$
& ARE $\downarrow$
& RPE\textsubscript{t} $\downarrow$
& RPE\textsubscript{r} $\downarrow$
& Acc. $\downarrow$
& Comp. $\downarrow$
& CD $\downarrow$
& Params $\downarrow$
& Latency $\downarrow$
& Memory $\downarrow$ \\
& & & (m) & ($^\circ$) & (m) & ($^\circ$)
& (m) & (m) & (m) & (M) & (s) & (GiB) \\
\midrule

\textbf{VGGT-$\Omega$}
& 30 & 48
& 0.31 & 9.99 & 0.18 & 0.93
& 1.15 & 1.08 & 1.11
& 605 & 1.98 & 8.2 \\
\rowcolor{oursblue}
$+$ \textbf{Ours}
& 30 & \textbf{30}
& 0.33 & 9.07 & 0.19 & 1.00
& 1.13 & 1.09 & 1.11
& 378\reduction{37.5}
& 1.58\reduction{20.4}
& 6.9\reduction{15.9} \\

\addlinespace
\textbf{VGGT-$\Omega$}
& 100 & 48
& 0.31 & 9.97 & 0.14 & 0.52
& 1.13 & 1.07 & 1.10
& 605 & 6.58 & 11.3 \\
\rowcolor{oursblue}
$+$ \textbf{Ours}
& 100 & \textbf{30}
& 0.32 & 8.51 & 0.14 & 0.54
& 1.12 & 1.08 & 1.10
& 378\reduction{37.5}
& 4.87\reduction{26.1}
& 10.0\reduction{11.3} \\

\addlinespace
\textbf{VGGT-$\Omega$}
& 200 & 48
& 0.31 & 10.02 & 0.14 & 0.45
& 1.13 & 1.07 & 1.10
& 605 & 12.16 & 13.2 \\
\rowcolor{oursblue}
$+$ \textbf{Ours}
& 200 & \textbf{30}
& 0.32 & 8.54 & 0.14 & 0.47
& 1.12 & 1.08 & 1.10
& 378\reduction{37.5}
& 8.67\reduction{28.7}
& 11.9\reduction{9.6} \\

\midrule

$\boldsymbol{\pi}^{\textbf{3}}$
& 30 & 36
& 0.36 & 10.59 & 0.21 & 0.96
& 1.25 & 1.12 & 1.18
& 454 & 1.16 & 6.5 \\
\rowcolor{oursblue}
$+$ \textbf{Ours}
& 30 & \textbf{20}
& 0.37 & 10.92 & 0.22 & 1.13
& 1.27 & 1.14 & 1.20
& 252\reduction{44.5}
& 0.80\reduction{30.9}
& 5.3\reduction{18.5} \\

\addlinespace
$\boldsymbol{\pi}^{\textbf{3}}$
& 100 & 36
& 0.35 & 9.99 & 0.16 & 0.54
& 1.23 & 1.10 & 1.17
& 454 & 4.70 & 8.0 \\
\rowcolor{oursblue}
$+$ \textbf{Ours}
& 100 & \textbf{20}
& 0.35 & 10.93 & 0.15 & 0.58
& 1.26 & 1.13 & 1.19
& 252\reduction{44.5}
& 2.99\reduction{36.4}
& 6.9\reduction{14.1} \\

\addlinespace
$\boldsymbol{\pi}^{\textbf{3}}$
& 200 & 36
& 0.34 & 10.01 & 0.16 & 0.49
& 1.23 & 1.10 & 1.17
& 454 & 9.40 & 9.1 \\
\rowcolor{oursblue}
$+$ \textbf{Ours}
& 200 & \textbf{20}
& 0.35 & 10.95 & 0.15 & 0.51
& 1.26 & 1.13 & 1.19
& 252\reduction{44.5}
& 5.78\reduction{38.5}
& 7.9\reduction{12.5} \\

\midrule

\textbf{VGGT}
& 30 & 48
& 0.90 & 9.90 & 0.52 & 1.24
& 1.13 & 1.12 & 1.12
& 605 & 1.78 & 8.8 \\
\rowcolor{oursblue}
$+$ \textbf{Ours}
& 30 & \textbf{38}
& 1.12 & 12.05 & 0.48 & 1.38
& 1.12 & 1.15 & 1.13
& 479\reduction{20.8}
& 1.48\reduction{14.2}
& 8.2\reduction{6.8} \\

\addlinespace
\textbf{VGGT}
& 100 & 48
& 0.86 & 10.03 & 0.40 & 0.70
& 1.12 & 1.10 & 1.11
& 605 & 6.68 & 10.9 \\
\rowcolor{oursblue}
$+$ \textbf{Ours}
& 100 & \textbf{38}
& 1.08 & 12.25 & 0.36 & 0.74
& 1.12 & 1.14 & 1.13
& 479\reduction{20.8}
& 5.62\reduction{15.8}
& 10.2\reduction{6.4} \\

\addlinespace
\textbf{VGGT}
& 200 & 48
& 0.86 & 10.04 & 0.40 & 0.62
& 1.12 & 1.10 & 1.11
& 605 & 13.21 & 12.5 \\
\rowcolor{oursblue}
$+$ \textbf{Ours}
& 200 & 38
& 1.08 & 12.26 & 0.36 & 0.65
& 1.12 & 1.14 & 1.13
& 479\reduction{20.8}
& 10.85\reduction{17.8}
& 11.9\reduction{5.5} \\

\bottomrule
\end{tabular}%
}
\vspace{-1em}
\end{table*}

\section{Raw Per-Metric Results}
\label{app:raw_results}

This section presents the raw per-metric counterparts to the figures in the main paper, without normalization or cross-metric averaging.
These results support the conclusions drawn from the normalized, aggregated results in the main text.

\begin{figure}[t]
    \centering
    \includegraphics[width=0.97\linewidth]
    {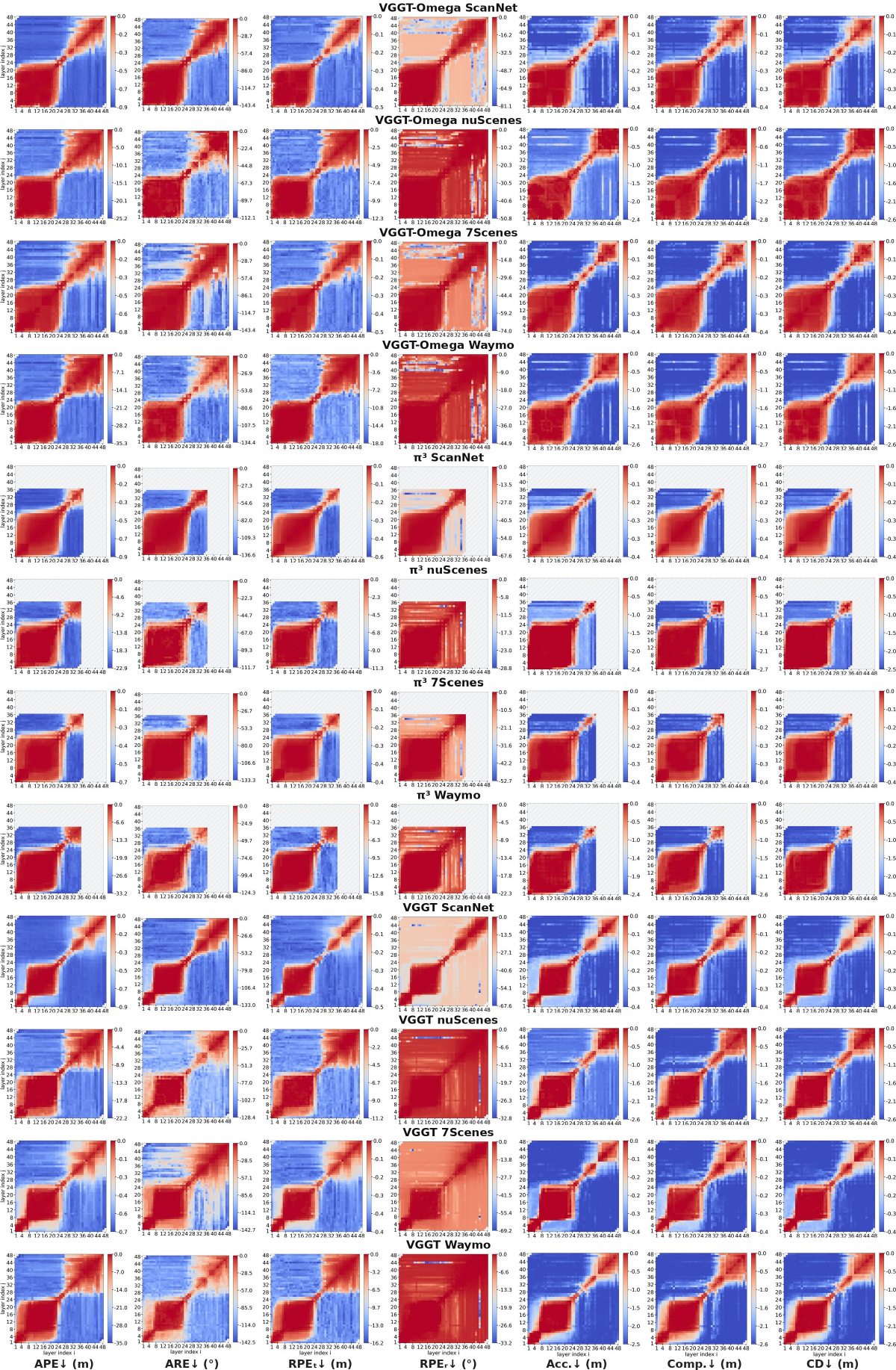}
    \caption{\textbf{A shared redundancy structure across the VGGT family.}
    Two separated low-degradation regions consistently appear across dataset domains and evaluation metrics, with the dominant region occurring early in the aggregator.
    Scores are mirrored across the diagonal.
    This is the unnormalized, per-metric counterpart to Fig.~\ref{fig:redundancy_heatmap}.
    }
\end{figure}

\begin{figure}[t]
    \centering
    \includegraphics[width=\linewidth]
    {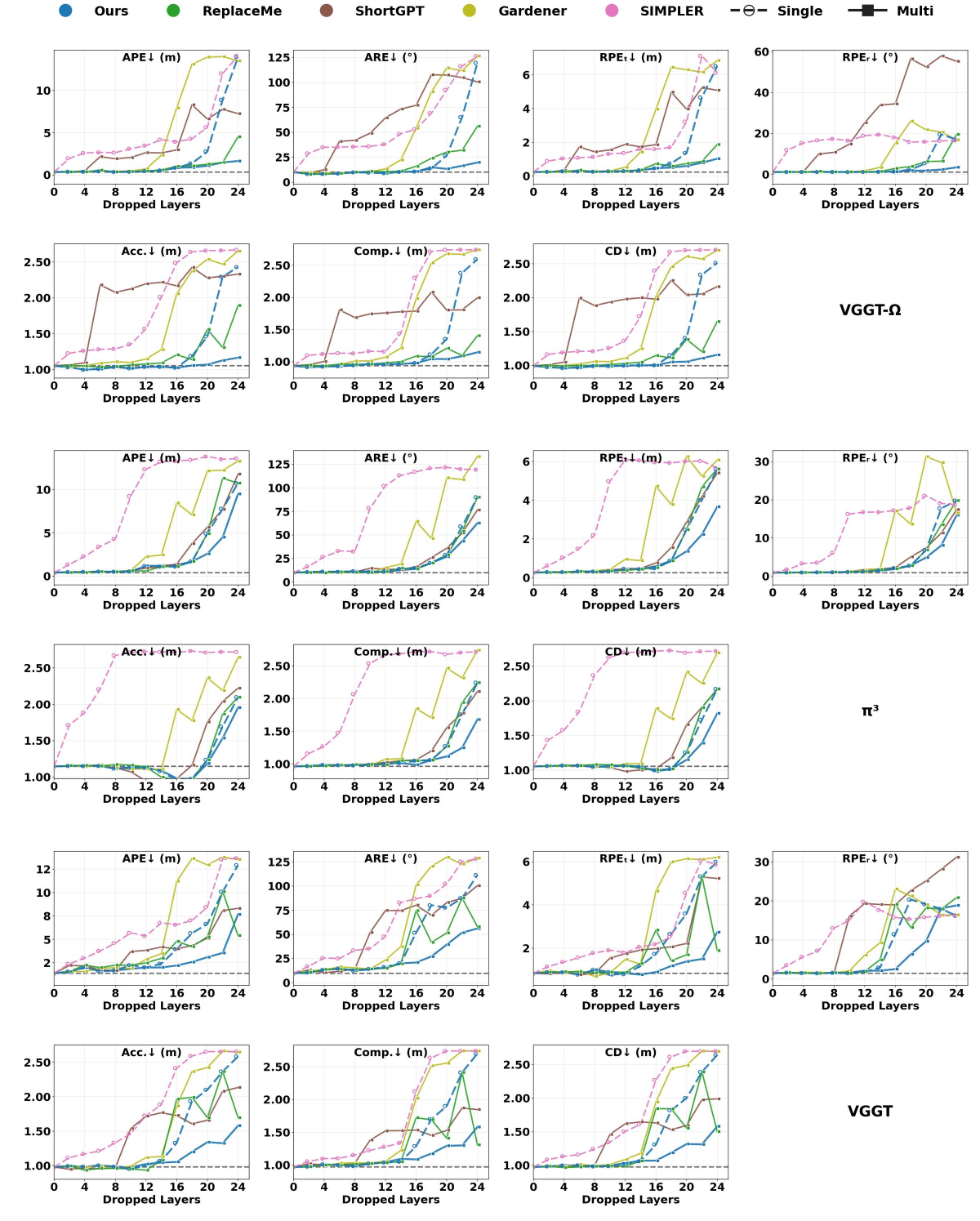}
    \caption{\textbf{Pruning quality and same-dataset generalization.}
    For each method and budget, one pruning configuration is selected based only on the combined 100-scene calibration set from ScanNet and nuScenes and applied unchanged to the held-out evaluation sets for same-dataset generalization.
    ``Single'' restricts removal to one contiguous interval, whereas ``Multi'' allows multiple intervals according to each method's search space; our method allows up to two separated intervals.
    This is the unnormalized, per-metric counterpart to Fig.~\ref{fig:pruning_curves}.
    }
\end{figure}

\begin{figure}[t]
    \centering
    \includegraphics[width=\linewidth]
    {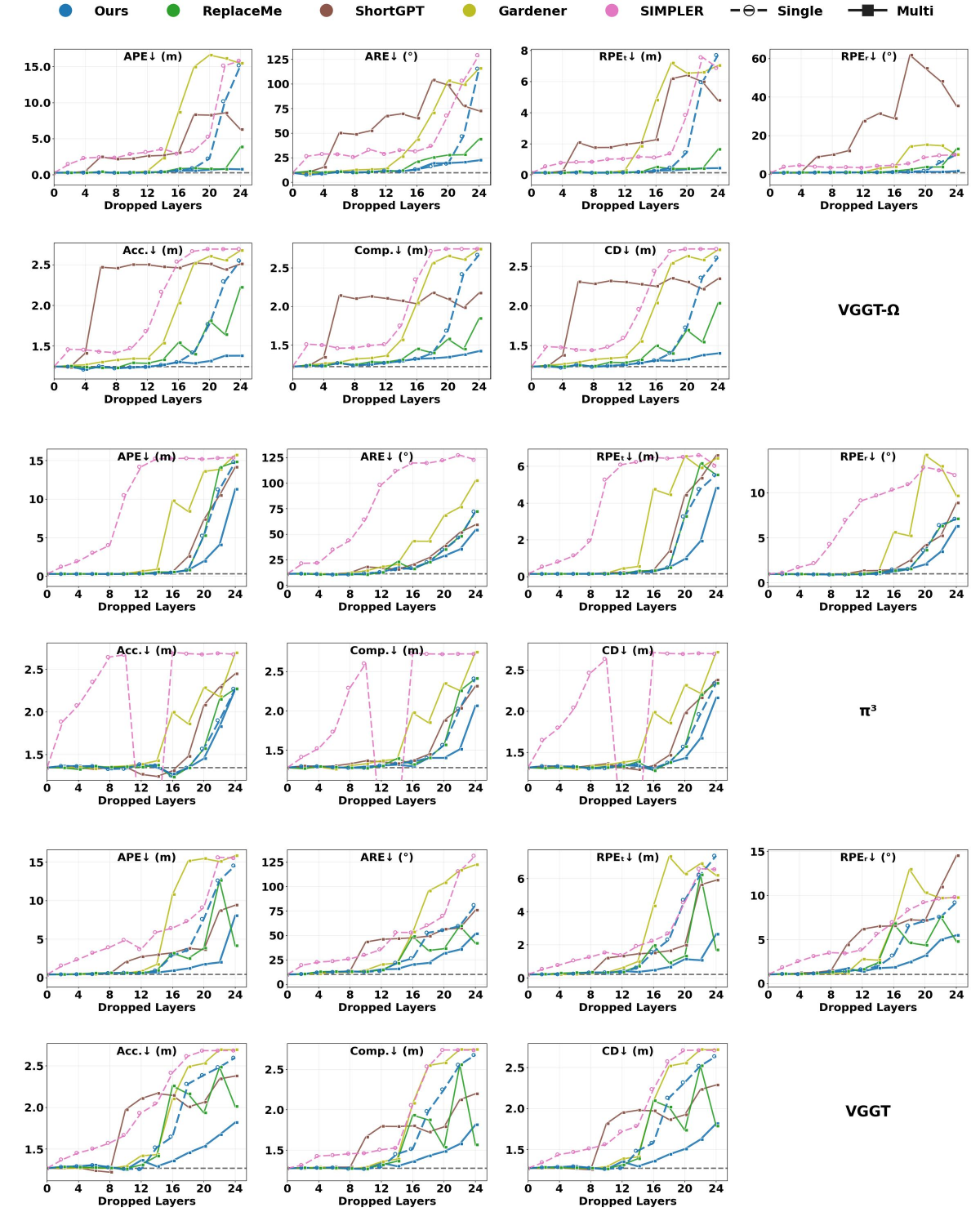}
    \caption{\textbf{Pruning quality and cross-dataset generalization.}
    For each method and budget, one pruning configuration is selected based only on the combined 100-scene calibration set from ScanNet and nuScenes and applied unchanged to the unseen datasets for cross-dataset generalization.
    ``Single'' restricts removal to one contiguous interval, whereas ``Multi'' allows multiple intervals according to each method's search space; our method allows up to two separated intervals.
    This is the unnormalized, per-metric counterpart to Fig.~\ref{fig:pruning_curves}.
    }
\end{figure}

\begin{figure}[t]
    \centering
    \includegraphics[width=\linewidth]
    {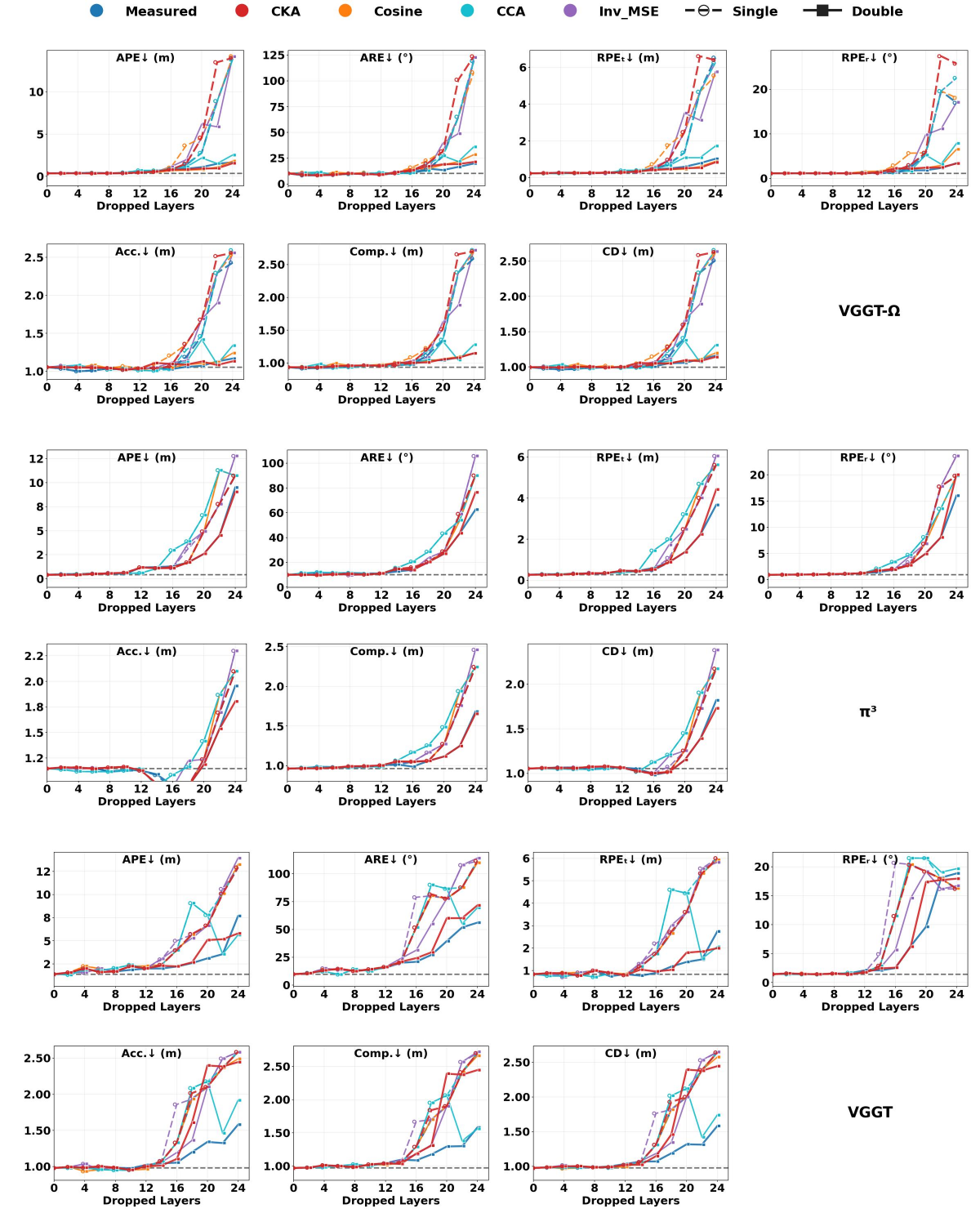}
    \caption{\textbf{Pruning quality and same-dataset generalization with representation-based proxies.}
    For each method and budget, one pruning configuration is selected based only on the combined 100-scene calibration set from ScanNet and nuScenes and applied unchanged to the held-out evaluation sets for same-dataset generalization.
    ``Single'' restricts removal to one contiguous interval, whereas ``Double'' allows up to two separated intervals.
    This is the unnormalized, per-metric counterpart to Fig.~\ref{fig:pruning_curves_sim}.
    }
\end{figure}

\begin{figure}[t]
    \centering
    \includegraphics[width=\linewidth]
    {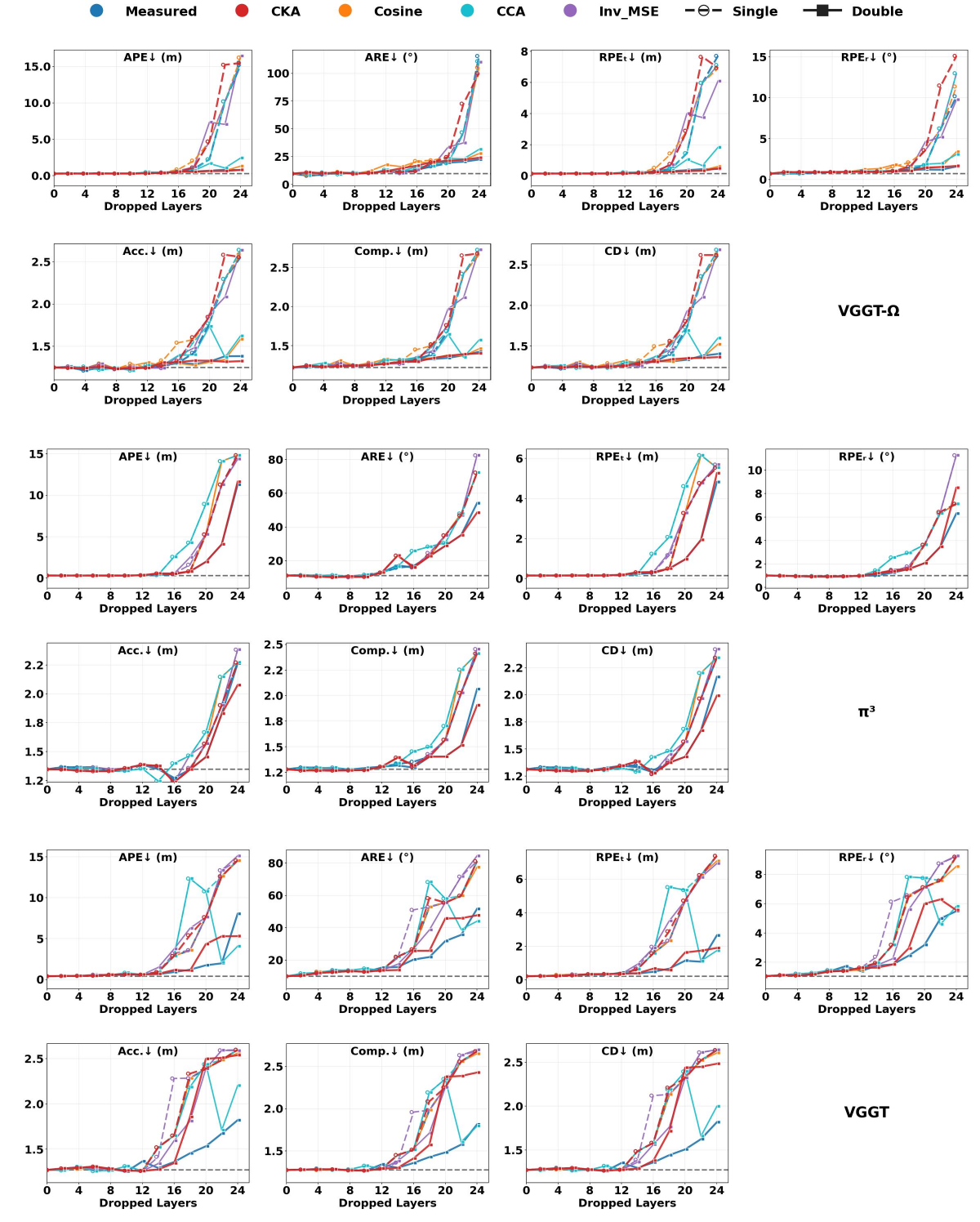}
    \caption{\textbf{Pruning quality and cross-dataset generalization with representation-based proxies.}
    For each method and budget, one pruning configuration is selected based only on the combined 100-scene calibration set from ScanNet and nuScenes and applied unchanged to the unseen datasets for cross-dataset generalization.
    ``Single'' restricts removal to one contiguous interval, whereas ``Double'' allows up to two separated intervals.
    This is the unnormalized, per-metric counterpart to Fig.~\ref{fig:pruning_curves_sim}.
    }
\end{figure}

\begin{figure}[t]
    \centering
    \includegraphics[width=\linewidth]
    {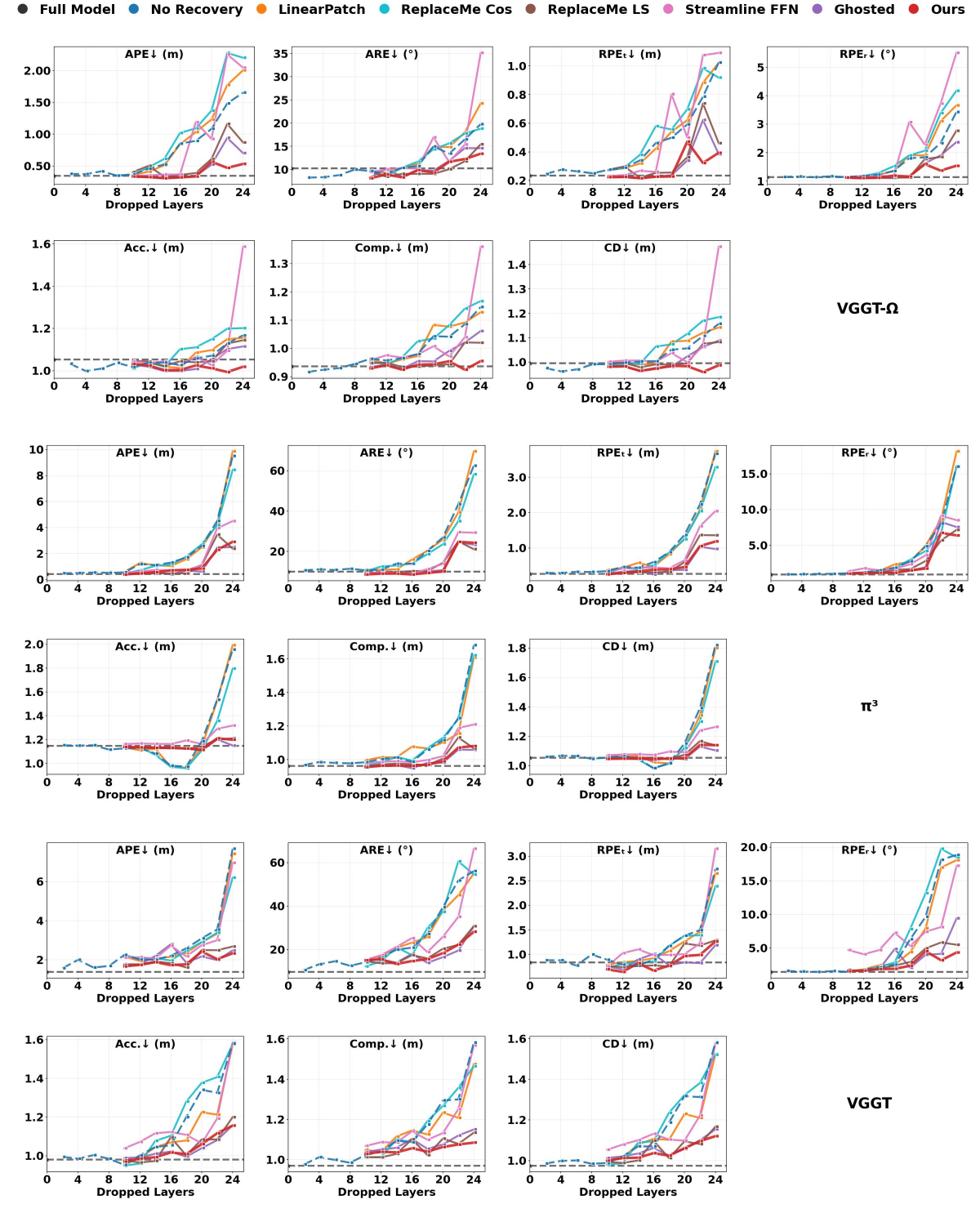}
    \caption{\textbf{Calibration-based post-pruning recovery benchmark.}
    Recovery transformations are fitted on the combined 100-scene calibration set from ScanNet and nuScenes and evaluated on their corresponding held-out evaluation sets.
    This is the unnormalized, per-metric counterpart to Fig.~\ref{fig:recovery_benchmark}.
    }
\end{figure}

\begin{figure}[t]
    \centering
    \includegraphics[width=\linewidth]
    {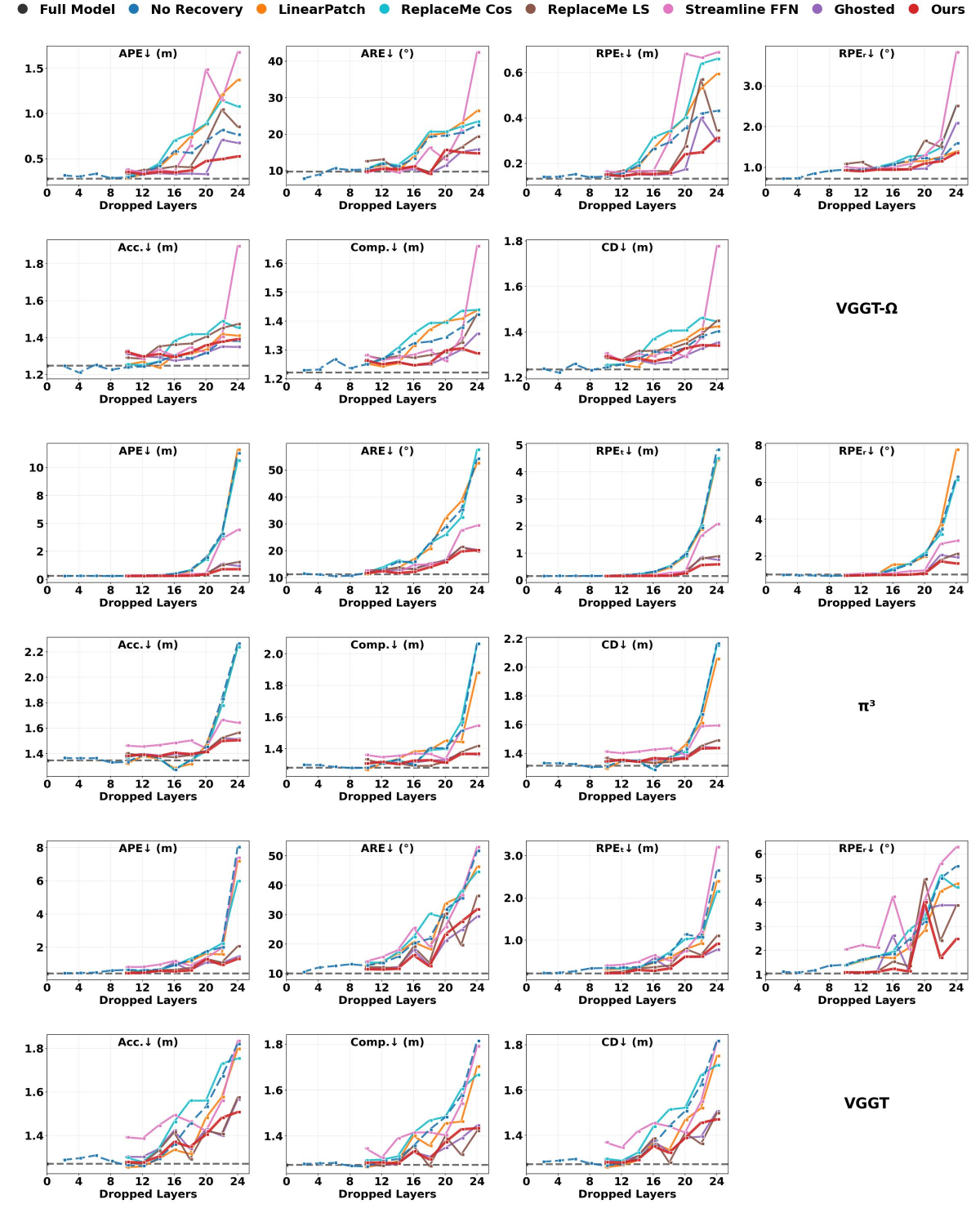}
    \caption{\textbf{Calibration-based post-pruning recovery benchmark.}
    Recovery transformations are fitted on the combined 100-scene calibration set from ScanNet and nuScenes and evaluated on the unseen 7Scenes and Waymo datasets.
    This is the unnormalized, per-metric counterpart to Fig.~\ref{fig:recovery_benchmark}.
    }
\end{figure}

\end{document}

%% file: math_commands.tex
\usepackage{amsmath,amsfonts,bm}

\def\eqref#1{equation~\ref{#1}}
\def\1{\bm{1}}

\DeclareMathAlphabet{\mathsfit}{\encodingdefault}{\sfdefault}{m}{sl}
\SetMathAlphabet{\mathsfit}{bold}{\encodingdefault}{\sfdefault}{bx}{n}